\documentclass[sn-nature]{sn-jnl}

\usepackage{graphicx}%
\usepackage{multirow}%
\usepackage{amsmath,amssymb,amsfonts}%
\usepackage{amsthm}%
\usepackage{mathrsfs}%
\usepackage[title]{appendix}%
\usepackage{xcolor}%
\usepackage{textcomp}%
\usepackage{manyfoot}%
\usepackage{booktabs}%
\usepackage{algorithm}%
\usepackage{algorithmicx}%
\usepackage{algpseudocode}%
\usepackage{listings}%
\usepackage{placeins}
\usepackage{booktabs}%
\usepackage{changes}

\theoremstyle{thmstyleone}%
\theoremstyle{thmstyletwo}%

\theoremstyle{thmstylethree}%

\begin{document}
	
	\title[DABench]{Benchmarking AI-based data assimilation to advance data-driven global weather forecasting}
	
	
	\author[1,2]{\fnm{Wuxin} \sur{Wang}}\email{wuxinwang@nudt.edu.cn}
	
	\author[2]{\fnm{Weicheng} \sur{Ni}}\email{niweicheng17@nudt.edu.cn}
	
	\author[3]{\fnm{Ben} \sur{Fei}}\email{benfei@cuhk.edu.hk}
	
	\author[3]{\fnm{Tao} \sur{Han}}\email{hantao10200@gmail.com}
	
	\author[1,2]{\fnm{Lilan} \sur{Huang}}\email{huanglilan18@nudt.edu.cn}
	
	\author[2]{\fnm{Taikang} \sur{Yuan}}\email{ytk@nudt.edu.cn}
	
	\author[2]{\fnm{Xiaoyong} \sur{Li}}\email{sayingxmu@nudt.edu.cn}
	
	\author*[2]{\fnm{Boheng} \sur{Duan}}\email{bhduan@nudt.edu.cn}
	
	\author*[3]{\fnm{Lei} \sur{Bai}}\email{bailei@pjlab.org.cn}
	
	
	\author*[1,2]{\fnm{Kaijun} \sur{Ren}}\email{renkaijun@nudt.edu.cn}
	
	\affil[1]{\orgdiv{College of Computer Science and Technology}, \orgname{National University of Defense Technology}, \orgaddress{\street{Deya Street}, \city{Changsha}, \postcode{410073}, \state{Hunan}, \country{China}}}
	
	\affil[2]{\orgdiv{College of Meteorology and Oceanography}, \orgname{National University of Defense Technology}, \orgaddress{\street{Deya Street}, \city{Changsha}, \postcode{410073}, \state{Hunan}, \country{China}}}
	
	\affil[3]{\orgname{Shanghai Artificial Intelligence Laboratory}, \orgaddress{\street{Longwen Road}, \postcode{200233}, \city{Shanghai}, \country{China}}}
	
	
	\abstract{
		Research on Artificial Intelligence (AI)-based Data Assimilation (DA) is expanding rapidly. However, the absence of an objective, comprehensive, and real-world benchmark hinders the fair comparison of diverse methods. Here, we introduce DABench, a benchmark designed for contributing to the development and evaluation of AI-based DA methods. By integrating real-world observations, DABench provides an objective and fair platform for validating long-term closed-loop DA cycles, supporting both deterministic and ensemble configurations. Furthermore, we assess the efficacy of AI-based DA in generating initial conditions for the advanced AI-based weather forecasting model to produce accurate medium-range global weather forecasting. Our dual-validation, utilizing both reanalysis data and independent radiosonde observations, demonstrates that AI-based DA achieves performance competitive with state-of-the-art AI-driven four-dimensional variational frameworks across both global weather DA and medium-range forecasting metrics. We invite the research community to utilize DABench to accelerate the advancement of AI-based DA for global weather forecasting.
	}

	
	
	
	\maketitle
	
	\section{Introduction}~\label{sec:intro}
	Understanding and predicting changes in the Earth system has long been a central challenge for science and society~\cite{gettelman2022future}. The rapid advancement of Artificial Intelligence (AI) has led to the emergence of several Large Weather Models (LWMs)~\cite{pathak2022fourcastnet,bi2023accurate,lam2023learning,chen2023fuxi,chen2023fengwu}. These models have demonstrated remarkable performance, achieving accuracy comparable to traditional Numerical Weather Prediction (NWP) systems, such as the Integrated Forecasting System (IFS) developed by the European Centre for Medium-Range Weather Forecasts (ECMWF)~\cite{bi2023accurate,lam2023learning,chen2023fuxi,chen2023fengwu}. However, these AI-based models currently rely on the analysis fields generated by traditional, computationally expensive NWP-based Data Assimilation (DA) systems as input, termed ``initial fields''. This dependence limits them from operating autonomously as self-contained data-driven systems capable of stable cycling forecasts~\cite{xiao2023fengwu,chen2023towards} and providing gridded weather forecasting products. 
	
	Recent progress in AI-based DA~\cite{tang2003enso,krasnopolsky2013application} spans several research directions, including DA function approximation on idealized scenarios~\cite{pawar2020long,fablet2021learning,wu2021fast,mccabe2021learning,fablet2023multimodal,wang2024four}, three-dimensional, multivariable initial condition estimation~\cite{chen2023towards,huang2024diffda,xu2024fuxi,wang2024accurate,sun2025data}, end-to-end weather forecasting~\cite{alexe2024graphdop,allen2025end}, and AI-driven four-dimensional variational (4DVar) DA frameworks~\cite{xiao2023fengwu,li2024fuxien4dvar,fan2026physically}. Despite the rapid progress, existing studies differ substantially in observational data, weather forecasting models, and evaluation metrics, which hinders objective comparison among different methods. Moreover, traditional DA benchmarks have primarily focused on theoretical mathematical problems using simulated data and simplified numerical models~\cite{vetra2018state,attia2019dates,grudzien2022dataassimilationbenchmarks,raanes2024dapper}, which cannot be directly applied to evaluating AI-based DA models for medium-range global weather forecasting. Consequently, there is an urgent need for a comprehensive benchmark that bridges the gap between AI-based DA models and real-world data-driven global weather forecasting.
	
	To further advance AI-based DA models and develop operational data-driven global weather forecasting systems, such a benchmark should evaluate several key features of the DA models: (1) the capacity to assimilate real-world observations for the long-term stable DA cycle, (2) the fidelity in generating analysis fields relative to real-world observations, and (3) the efficacy in initializing LWMs for skillful global weather forecasts.
	
	Here, we introduce DABench, an open-source benchmark specifically designed for AI-based DA in data-driven global weather forecasting. DABench integrates ERA5 reanalysis~\cite{hersbach2020era5} with quality-controlled real-world conventional observations collected from the Global Data Assimilation System (GDAS) prepbufr~\cite{cisl_rda_dsd337000}. In contrast to previous AI-based DA research that relies exclusively on ERA5 for validation, we incorporate independent radiosonde observations as an additional reference. This dual-validation approach enables a comprehensively assesses the accuracy of AI-based DA models in estimating complete atmospheric fields comparable to ERA5 while simultaneously verifying their ability to generate analysis fields that closely resemble real-world observations. By employing Pangu-Weather~\cite{bi2023accurate} as the representative weather forecasting model, we conducted a fair evaluation of several open-source AI-based DA models~\cite{fablet2021learning,yasuda2023spatio,chen2023adas,wang2024accurate,manshausen2025generative} and the recently proposed AI-driven 4DVar framework~\cite{fan2026physically} under both deterministic and ensemble configurations. Our results demonstrate that a data-driven global weather forecasting system, incorporating AI-based DA models, can sustain a stable one-year DA cycle without notable error accumulation or systematic drift. Furthermore, the evaluation reveals that AI-based DA achieves performance metrics in both DA cycles and medium-range global weather forecasting that are competitive with the advanced AI-driven 4DVar framework~\cite{fan2026physically}, highlighting the potential of AI-based DA models for real-world applications.
	
	Overall, our contributions can be summarized as follows:
	\begin{itemize}
		\item \textbf{Comprehensive Dataset:} The core of DABench is a benchmark comprising quality-controlled real-world observations for the DA model training and comprehensive performance evaluation. The dataset details are provided in Section~\ref{sec:4.2}.
		\item \textbf{Standardized Evaluation Metrics:} We provide standardized metrics for model comparison. Our benchmark assesses the quality of analysis fields during the one-year deterministic and Ensemble DA (EDA) cycle, as well as the performance of medium-range global weather forecasting initialized by these fields.
		\item \textbf{Objectively evaluate AI-based DA:} We provide an objective evaluation of various open-source AI-based DA models and the advanced AI-driven 4DVar method using this benchmark. This evaluation identifies specific performance gaps and areas where AI-based DA requires further optimization.
		\item \textbf{Validation of AI-based DA Potential:} By incorporating both ERA5 and independent radiosonde observations as references, we demonstrate that AI-based DA can match the performance of the 4DVar method across global weather DA and medium-range forecasting metrics.
	\end{itemize}
	
	\section{Results}~\label{sec:results}
	
	\subsection{The motivation and framework of DABench} \label{sec2.1}
	\begin{figure}[htbp]
		\centering
		\noindent\includegraphics[width=0.95\textwidth]{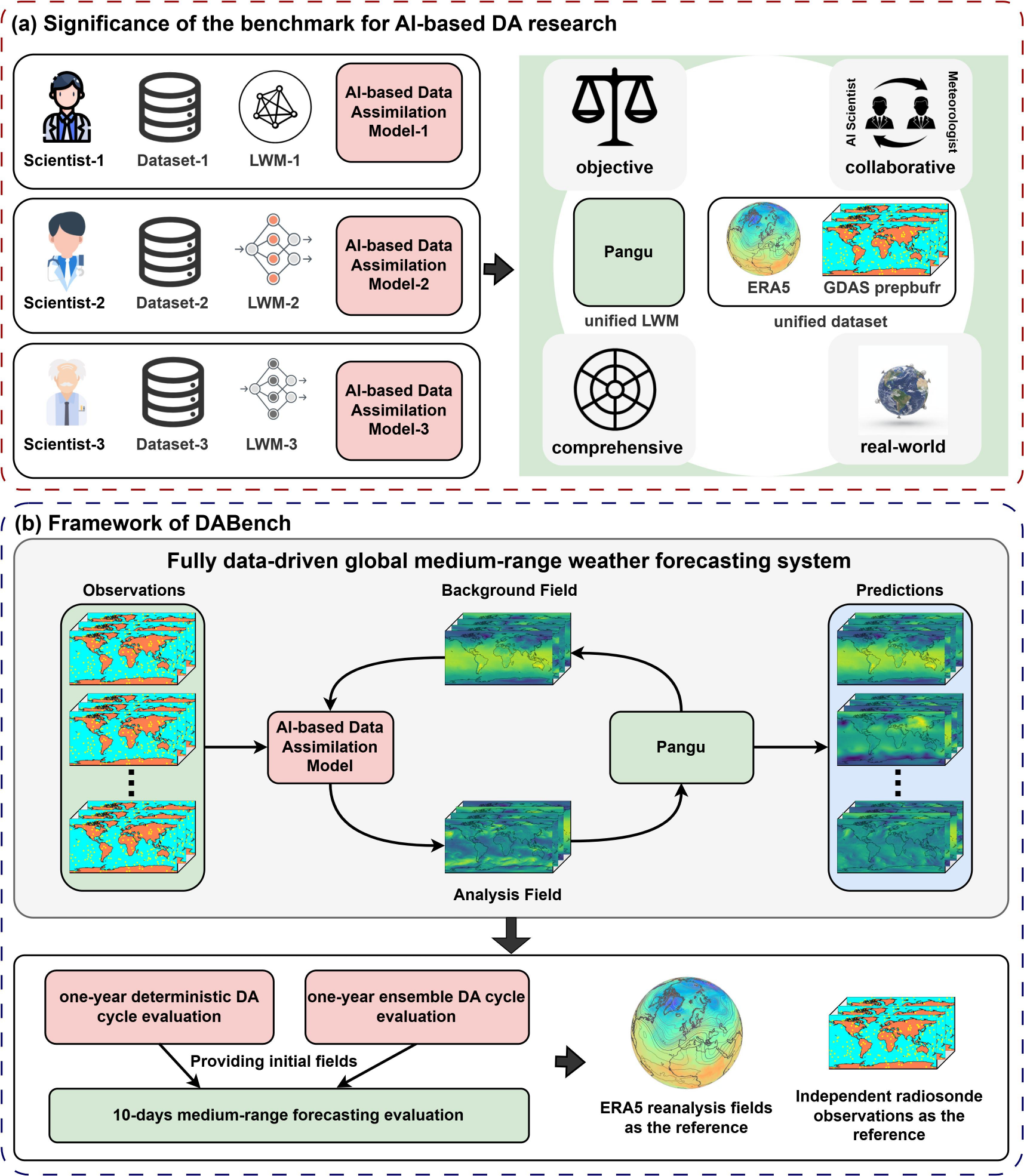}
		\caption{\textbf{Overview of DABench.} (a) Schematic illustration highlighting the significance of the benchmark for AI-based DA research. (b) Framework of DABench. The data-driven medium-range global weather forecasting system consists of two core components: the forecasting component, which utilizes the Pangu-Weather~\cite{bi2023accurate} model for generating forecasts, and the DA component, which integrates the background field and observations using the DA baselines evaluated in this study to produce the analysis required for initializing the forecasting task. The system is developed and evaluated using both ERA5 reanalysis and independent radiosonde observations, as well as deterministic and ensemble DA cycle configurations. The DA models are trained using real-world observations to approximate the ERA5 reanalysis, and are then evaluated against ERA5 for overall performance and against independent radiosonde observations for their ability to estimate the real atmosphere. Finally, medium-range weather forecasting is assessed using their outputs as initial fields to evaluate their potential for operational applications.}~\label{fig1}
	\end{figure}

	Figure~\ref{fig1} (a) illustrates the motivation behind DABench. Current AI-based DA research suffers from fragmentation, with researchers independently collecting observational data and developing disparate forecasting and DA models~\cite{fablet2021learning,yasuda2023spatio,chen2023adas,wang2024accurate,manshausen2025generative}. This fragmentation hinders objective comparison and reproducibility. DABench addresses this challenge by providing an objective, collaborative, and comprehensive evaluation platform based on real-world observations. By fostering synergy between AI scientists and meteorologists, DABench can accelerate the development of operational data-driven medium-range global weather forecasting systems. 
	
	Figure~\ref{fig1} (b) presents the framework of a data-driven global weather forecasting system. The forecasting component generates predictions with lead times from 6 to 240 hours, while the DA component integrates observations to update weather states, enabling stable and continuous cycling forecasts. We establish baselines using three categories of methods: (1) a minimalist DA model based on SwinTransformer~\cite{liu2021swin}, a common architecture in LWMs, and (2) several open-source SOTA AI-based DA methods from recent literature, including 4DVarNet~\cite{fablet2021learning}, 4DSRDA~\cite{yasuda2023spatio}, Adas~\cite{chen2023adas}, SDA~\cite{manshausen2025generative}, and 4DVarFormer~\cite{wang2024accurate}, and (3) a physically consistent Latent 4DVar (L4DVar)~\cite{fan2026physically} framework that leverages the automatic differentiation of Pangu-Weather to replace the tangent-linear and adjoint models. Detailed descriptions of these baselines are provided in Section~\ref{sec:4.4}.
	
	The DA cycle evaluation assesses the performance of 12-hourly analysis fields over a one-year period to determine the models' ability to integrate time-varying background fields and observations. The initial background field is derived from a 120-hour prediction initialized with ERA5 reanalysis at 00:00 UTC on December 27, 2022, with the DA cycle commencing at 00:00 UTC on January 1, 2023. Each 12-hour DA window (DAW) involves assimilating observations throughout the window and performing a 12-hour prediction to generate the background field for the subsequent DAW. Performance metrics include Weighted Root Mean Square Error (WRMSE), Weighted Bias (WBias), and power spectra for the deterministic configuration, with Continuous Ranked Probability Score (CRPS) and the Spread-Skill Ratio (SSR) for the Ensemble DA (EDA) configuration. To rigorously assess real-world applicability, we evaluate against independent radiosondes from GDAS prepbufr using Observational Root Mean Square Error (ORMSE) and the Observational Bias (OBias) metrics.
	
	To evaluate medium-range weather forecasting performance, we assess predictions initialized from the analysis fields sampled at 12-hour intervals. Specifically, we compute the WRMSE, ORMSE, Anomaly Correlation Coefficient (ACC), and Activity metrics for forecasts with lead times up to 10 days. Detailed information on all evaluation metrics is presented in Section~\ref{sec:4.6}. Variable abbreviations follow the conventions established in Table~\ref{tab1}. For pressure level variables, abbreviations follow a standardized format: the variable identifier is appended with the specific pressure level (e.g., Z500).
	
	\subsection{Results of the one-year DA cycle} \label{sec2.2}
	\subsubsection{Evaluation using ERA5 as the reference} \label{sec2.2.1}
	ECMWF has operationalized EDA in its NWP system to generate initial fields~\cite{EDA_EC}. Consequently, evaluating the performance of DA baselines in the context of EDA is of critical importance. To the best of our knowledge, this topic has been addressed in only a limited number of prior studies. In this section, all DA baselines are configured with one control member and ten perturbation members, resulting in 11 ensemble members. For further details regarding the EDA methods employed in this study, please refer to Section~\ref{sec:4.5}. Here, we test the EDA using both ERA5 and independent radiosondes as references, respectively.
	
	\begin{figure}[htbp]%
		\centering
		\includegraphics[width=0.95\textwidth]{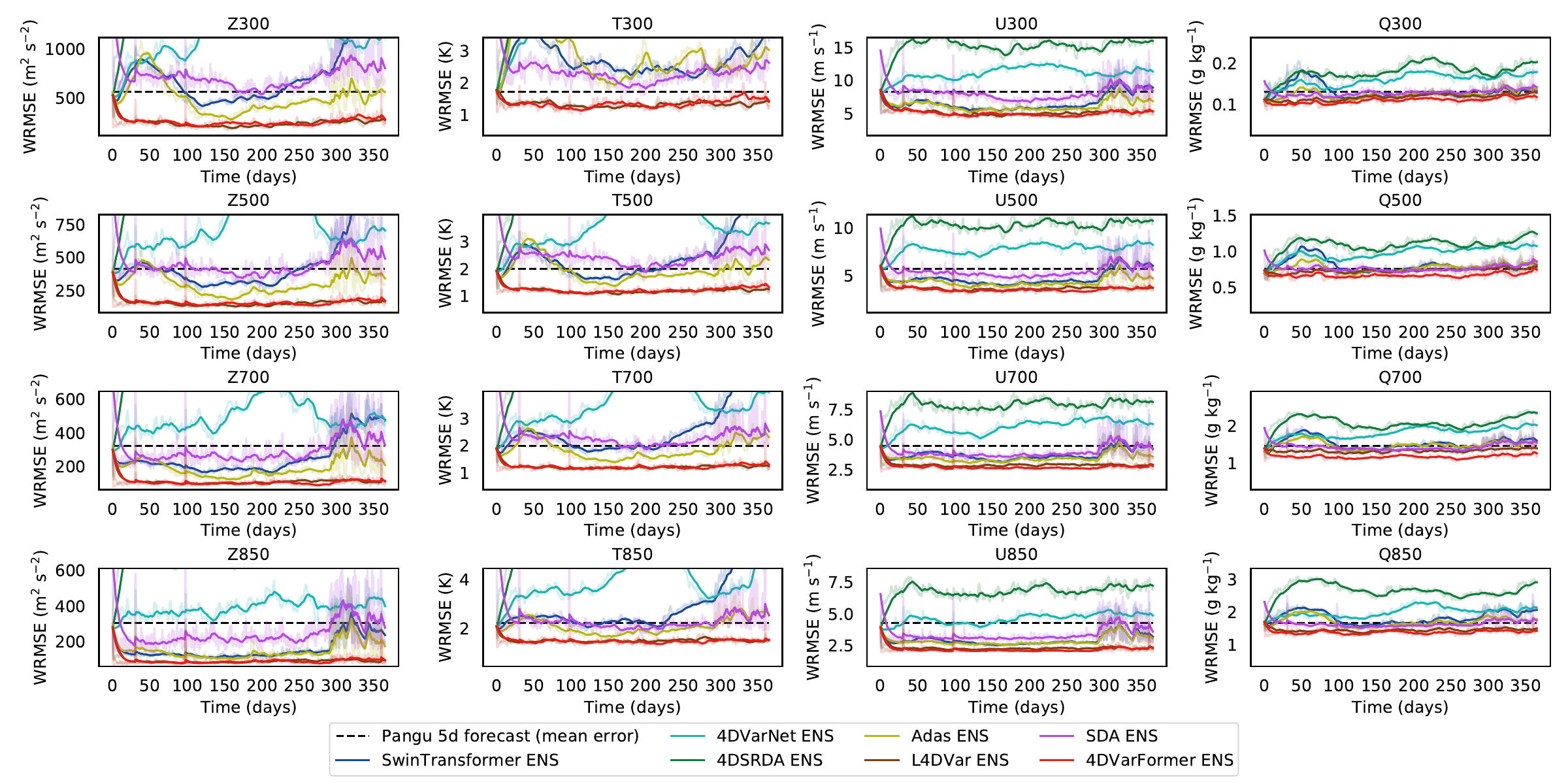}
		\caption{\textbf{WRMSE of baselines over a one-year DA cycle using ERA5 as reference.} The 5-day Pangu forecast is depicted by a black dashed line. The AI-based DA results are color-coded as follows: SwinTransformer (dark blue), 4DVarNet (light blue), 4DSRDA (dark yellow-green), Adas (light yellow-green), L4DVar (brown), SDA (purple), and 4DVarFormer (red). Evaluations were performed at 00:00 and 12:00 UTC daily throughout the year. Each subplot corresponds to a distinct variable, as indicated by the title.}
		\label{fig2}
	\end{figure}
	\FloatBarrier
	
\begin{figure}[htbp]%
		\centering
		\includegraphics[width=0.95\textwidth]{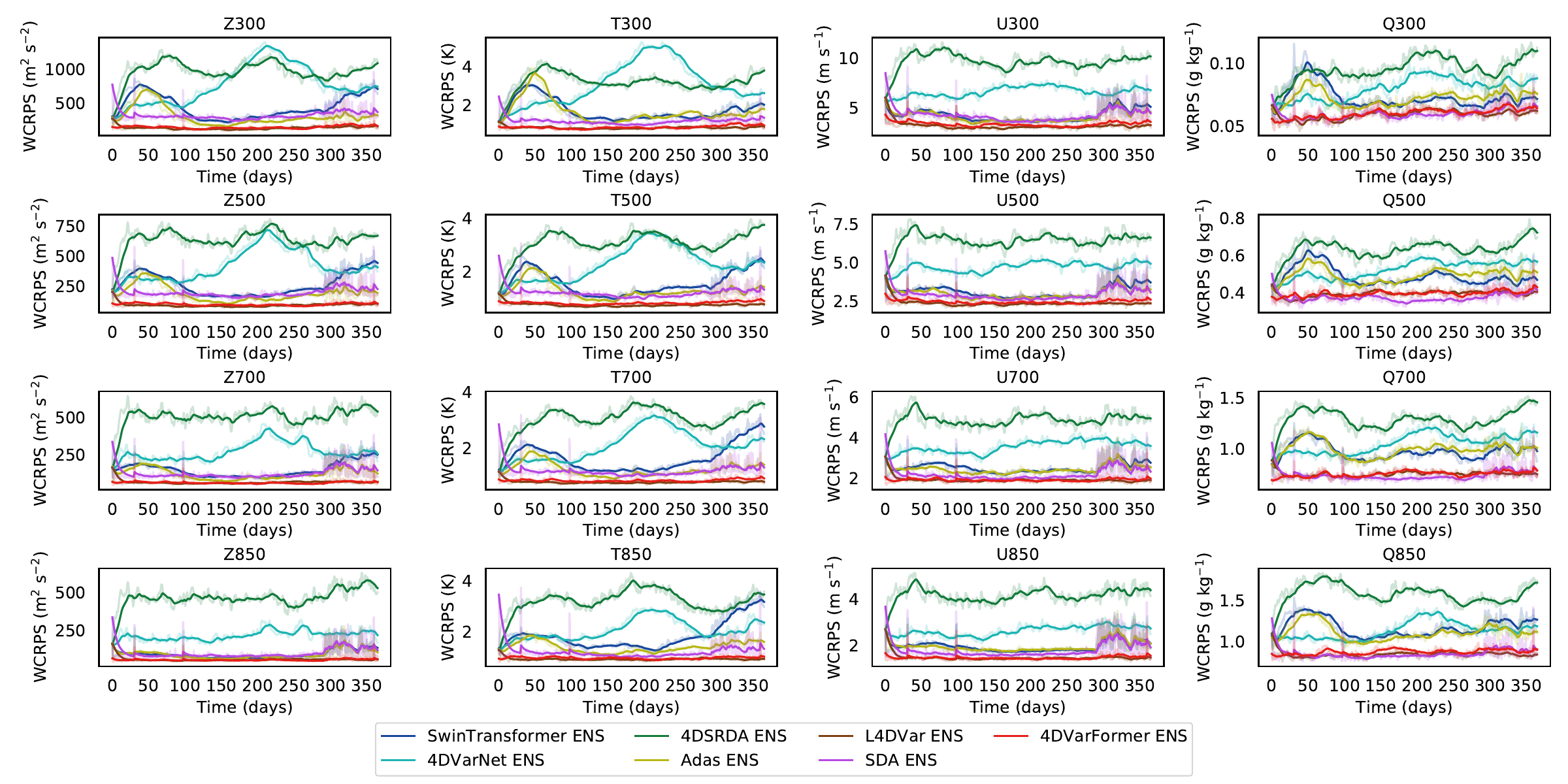}
		\caption{\textbf{CRPS of baselines over a one-year DA cycle using ERA5 as reference.} The AI-based DA results are color-coded as follows: SwinTransformer (dark blue), 4DVarNet (light blue), 4DSRDA (dark yellow-green), Adas (light yellow-green), L4DVar (brown), SDA (purple), and 4DVarFormer (red). Evaluations were performed at 00:00 and 12:00 UTC daily throughout the year. Each subplot corresponds to a distinct variable, as indicated by the title.}
		\label{fig3}
	\end{figure}

	Figure~\ref{fig2} and Figure~\ref{fig3} show the annual WRMSE and CRPS of the analysis fields generated by ensemble DA cycles using ERA5 as the reference. Each subplot corresponds to a variable and skill (y-axis) is shown at 12-hour steps over 365-day horizons (x-axis). The original data is represented with reduced opacity to enhance clarity, while solid lines depict values smoothed using an Exponential Moving Average (EMA) with a 15-point window. Notably, both 4DVarNet and 4DSRDA struggle to sustain stable DA cycles over extended periods. In contrast, while SwinTransformer, Adas, and SDA demonstrate relatively stable WRMSE performance, they do not surpass the 5-day forecast accuracy of the Pangu-Weather model across all variables and all times. Conversely, 4DVarFormer exhibits superior stability and achieves WRMSE and CRPS comparable to L4DVar, indicating that incorporating 4DVar constraints into AI-based models can substantially enhance DA accuracy. Notably, even without an explicit background error covariance matrix to enforce physical consistency, 4DVarFormer effectively captures global atmospheric characteristics, enabling efficient DA and forecast correction. Furthermore, a comparison with deterministic DA results (see Supplementary Text, Figure S3, as well as Tables S7 and S13) reveals that the WRMSE for each baseline decreases under the EDA configuration. This indicates that incorporating ensemble-based uncertainty estimation allows AI-based DA models to better capture atmospheric characteristics, thereby enhancing DA accuracy.
	
	\FloatBarrier
	
\begin{figure}[htbp]%
		\centering
		\includegraphics[width=0.95\textwidth]{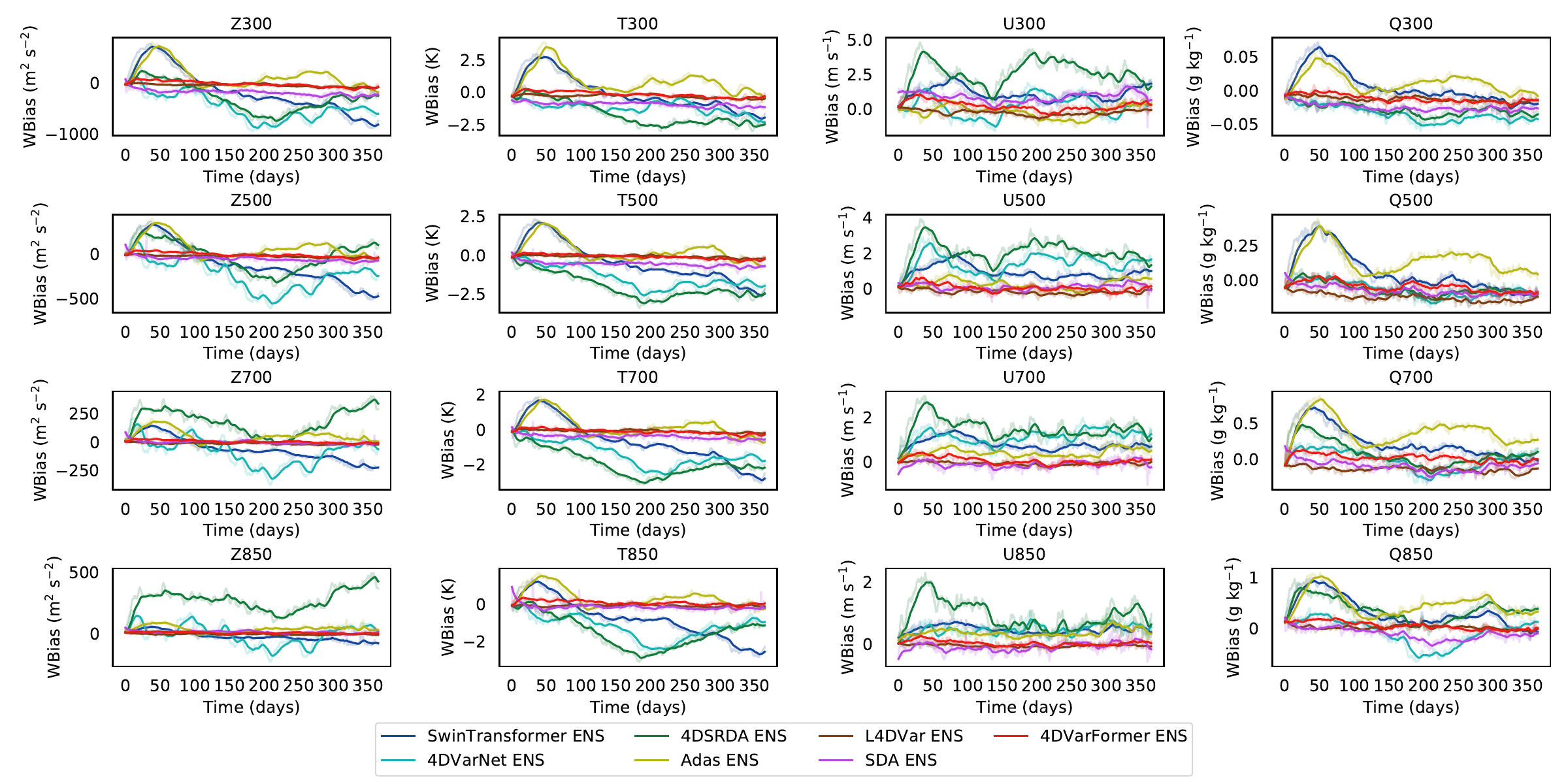}
		\caption{\textbf{WBias of baselines over a one-year DA cycle using ERA5 as reference.} The AI-based DA results are color-coded as follows: SwinTransformer (dark blue), 4DVarNet (light blue), 4DSRDA (dark yellow-green), Adas (light yellow-green), L4DVar (brown), SDA (purple), and 4DVarFormer (red). Evaluations were performed at 00:00 and 12:00 UTC daily throughout the year. Each subplot corresponds to a distinct variable, as indicated by the title.}
		\label{fig4}
	\end{figure}

	Figure~\ref{fig4} illustrates the temporal evolution of the WBias for analysis fields generated by DA cycling, with ERA5 serving as the reference. The observed instability in WBias suggests that SwinTransformer, 4DVarNet, and 4DSRDA struggle to generalize to complex, evolving background field distributions when assimilating sparse observations. This limitation underscores the necessity of mitigating error accumulation arising from the DA and forecasting cycle. Adas exhibits lower WBias than SwinTransformer for most variables, indicating that incorporating observational uncertainty effectively reduces analysis instability. SDA, which learns the ERA5 prior distribution through pre-training, maintains relatively low bias, despite estimating the complete atmospheric state solely from observations. Among all baselines, L4DVar and 4DVarFormer achieve WBias values closest to zero, demonstrating the superior stability of AI-driven 4DVar frameworks. 
	
	\FloatBarrier
	
	For detailed evaluation metric values of DA cycle experiments, please see the Supplementary Text and Tables S7 to S10. For DA performance results under the deterministic DA configuration, please refer to the Supplementary Text, Figures S3, S4, and S7, as well as Tables S13 and S14.
	
	\subsubsection{Evaluation using independent radiosondes as the reference} \label{sec2.2.2}
	
\begin{figure}[htbp]%
		\centering
		\includegraphics[width=0.95\textwidth]{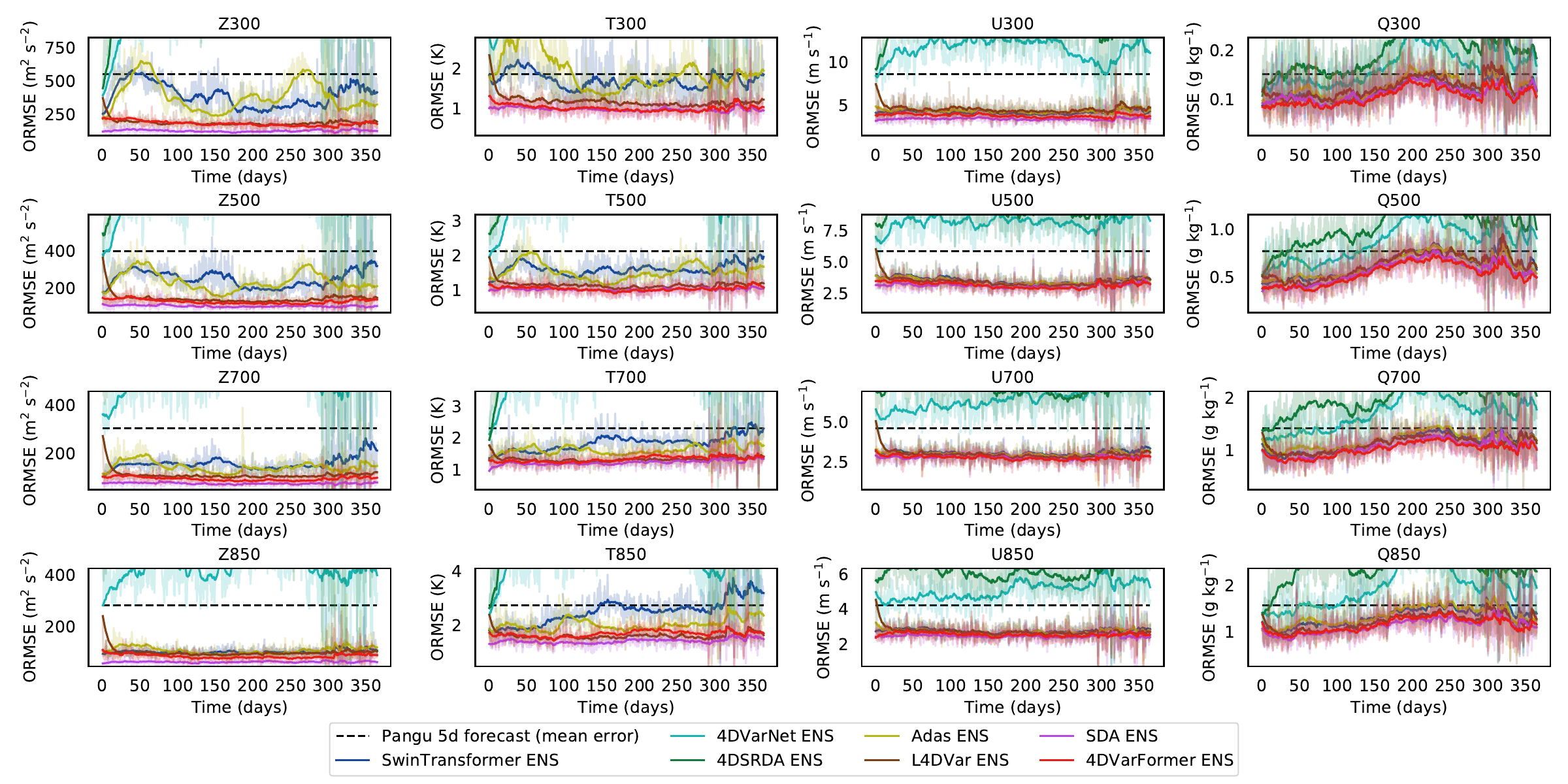}
		\caption{\textbf{ORMSE of baselines over a one-year DA cycle using independent sounding as reference.} The 5-day Pangu forecast is depicted by a black dashed line. The AI-based DA results are color-coded as follows: SwinTransformer (dark blue), 4DVarNet (light blue), 4DSRDA (dark yellow-green), Adas (light yellow-green), L4DVar (brown), SDA (purple), and 4DVarFormer (red). Evaluations were performed at 00:00 and 12:00 UTC daily throughout the year. Each subplot corresponds to a distinct variable, as indicated by the title.}
		\label{fig5}
	\end{figure}

	Figure~\ref{fig5} presents the annual ORMSE of analysis fields generated by baseline DA methods in the EDA configuration for key variables, using independent radiosondes as the reference. Each subplot corresponds to a specific variable, with ORMSE values (y-axis) plotted at 12-hour intervals over a 365-day horizon (x-axis). Raw data are displayed with reduced opacity to enhance clarity, while solid lines represent values smoothed using a 15-point EMA. When evaluated against independent radiosonde observations, 4DVarNet, 4DSRDA, SwinTransformer, Adas, and L4DVar show minimal improvement relative to their deterministic DA configurations. In contrast, SDA and 4DVarFormer achieved performance gains in the EDA configuration compared to the deterministic setting. Notably, although SDA achieved the lowest ORMSE against independent radiosondes in ensemble settings, its capacity to represent the global atmospheric field remains limited, highlighting its difficulty in effectively propagating observed information spatially. Furthermore, 4DVarFormer outperformed L4DVar in ORMSE for nearly all upper-level variables. This indicates that AI-based DA models can achieve performance comparable to or superior to 4DVar, thereby demonstrating their viability for operational implementation. 
	
	\FloatBarrier
	
	For more detailed results, please refer to the Supplementary Text, Figure S1, and Tables S11 and S12. For DA performance results under the deterministic DA configuration, please refer to the Supplementary Text, Figures S5 and S6, as well as Tables S15 and S16.
	
	\subsection{Results of the 10-day medium-range weather forecasting} \label{sec2.3}
	
	For DA models intended to support a fully data-driven global weather-forecasting system, assessment must encompass the skill of medium-range weather forecasts that are produced when the unified forecast model is initialized with the corresponding analyses. Accordingly, we examined the 10-day medium-range weather forecast performance of Pangu-Weather when initialized by analysis fields obtained from the EDA configuration.
	
\begin{figure}[htbp]%
		\centering
		\includegraphics[width=\textwidth]{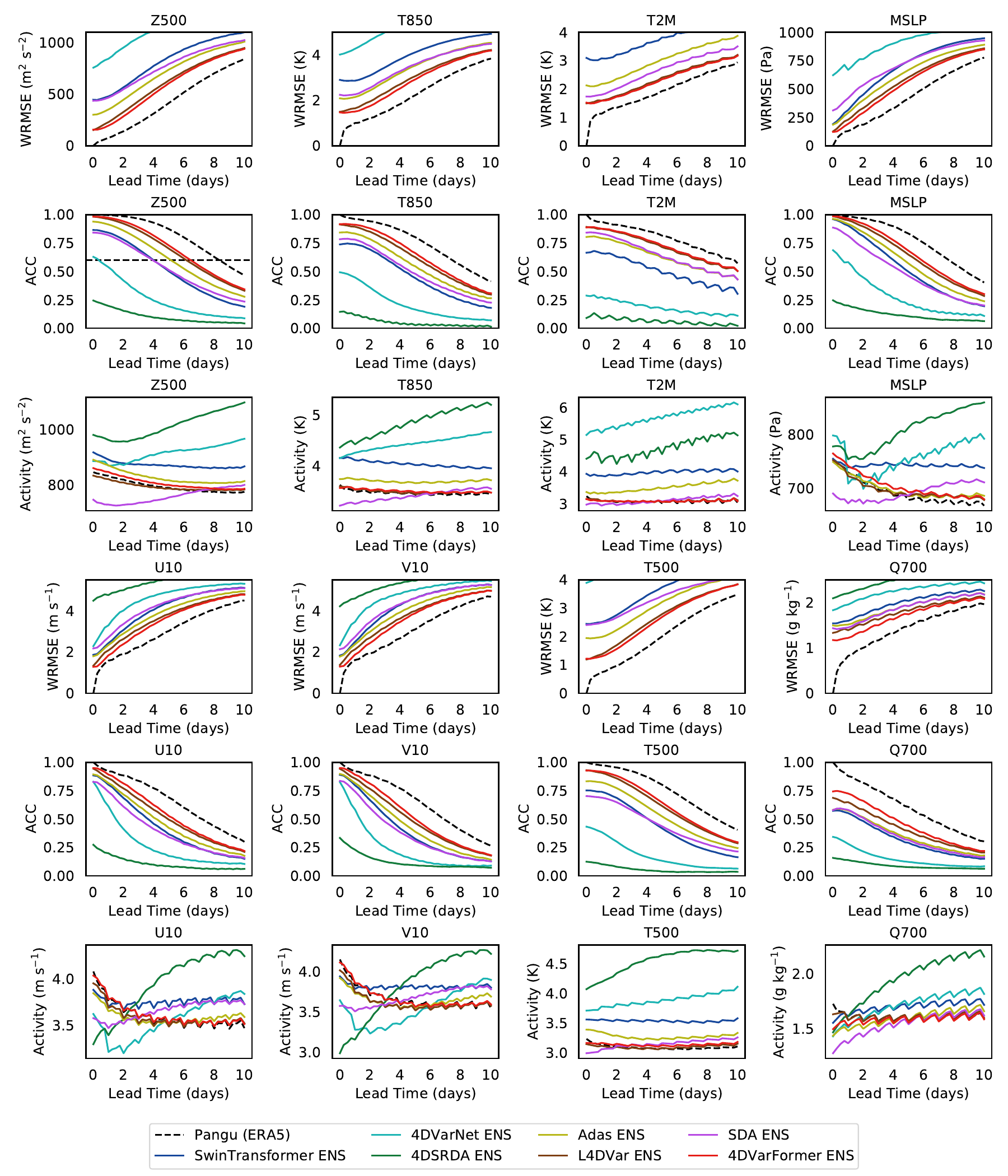}
		\caption{\textbf{WRMSE, ACC, and Activity metrics of baselines for 10-day medium-range predictions initialized by the analysis fields generated by EDA.} The Pangu forecast using ERA5 as the initial field is depicted by a black dashed line. The medium-range weather forecasting results initialized by AI-based DA are color-coded as follows: SwinTransformer (dark blue), 4DVarNet (light blue), 4DSRDA (dark yellow-green), Adas (light yellow-green), L4DVar (brown), SDA (purple), and 4DVarFormer (red). These calculations are done for each day of the year at 00:00 UTC and 12:00 UTC. All metrics are computed against ERA5. Each subplot represents a single variable, as indicated in the subplot titles.}
		\label{fig6}
	\end{figure}

	Figure \ref{fig6} illustrates the performance of the medium-range weather forecasting initialized by the analysis fields obtained from assimilating GDAS prepbufr observations. The results indicate that only SwinTransformer, Adas, L4DVar, SDA, and 4DVarFormer can produce valid medium-range weather forecasts (with an ACC greater than 0.6 for Z500). Notably, the skillful lead time of Pangu-Weather initialized by 4DVarFormer can exceed all other baselines. This finding underscores the significant potential of 4DVarFormer for operational applications, particularly when combined with Pangu in developing a fully data-driven global weather forecasting system. Despite these advancements, current AI-based DA methods still struggle to outperform the performance of forecasts initialized with ERA5 reanalysis. This gap suggests that AI-driven DA models require further refinement. It also underscores that assimilating conventional observations alone remains insufficient for optimal accuracy. Furthermore, most existing models rely on supervised learning, which inherently constrains their performance to the quality of the training labels. Consequently, investigating novel training paradigms for assimilation models is a critical frontier for future research.
	
	The Activity metric quantifies the extent to which extreme values are represented within the forecast field. As illustrated in Figure~\ref{fig6}, the Activity values for forecasts initialized by 4DVarFormer and L4DVar analysis fields exhibit the strongest resemblance to those driven by ERA5. This similarity suggests that both 4DVarFormer and L4DVar, by leveraging Pangu-derived adjoint constraints, generate initial conditions that are highly compatible with the Pangu-Weather model, thereby preserving physical consistency throughout the forecast evolution. In contrast, SwinTransformer, 4DVarNet, 4DSRDA, and Adas generally exhibit higher Activity values, suggesting that the initial fields generated by these models contain small-scale noise that progressively accumulates during forecasting. Conversely, SDA displays lower Activity in the early forecast stages, which subsequently increases to exceed the levels observed in ERA5-driven forecasts. This pattern indicates that while the ensemble SDA produces relatively smooth analysis fields, imbalances in the initial conditions eventually induce extreme value noise as the forecast evolves.
	
\begin{figure}[htbp]%
		\centering
		\includegraphics[width=\textwidth]{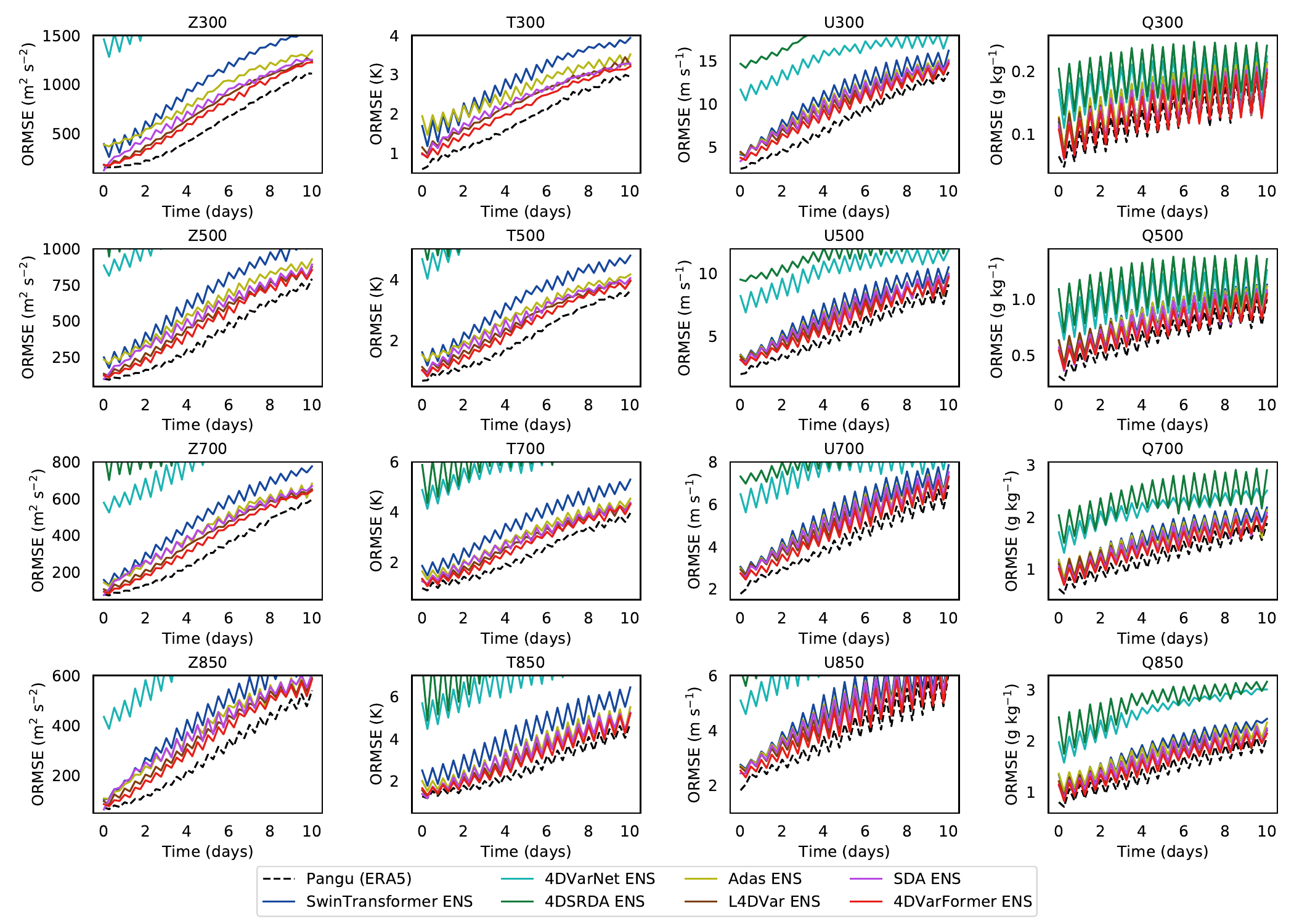}
		\caption{\textbf{ORMSE of baselines for 10-day medium-range predictions initialized by the analysis fields generated by EDA using independent sounding as reference.} The Pangu forecast using ERA5 as the initial field is depicted by a black dashed line. The medium-range weather forecasting results initialized by AI-based DA are color-coded as follows: SwinTransformer (dark blue), 4DVarNet (light blue), 4DSRDA (dark yellow-green), Adas (light yellow-green), L4DVar (brown), SDA (purple), and 4DVarFormer (red). These calculations are done for each day of the year at 00:00 UTC and 12:00 UTC. All metrics are computed against ERA5. Each subplot represents a single variable, as indicated in the subplot titles.}
		\label{fig7}
	\end{figure}

	We also evaluated the forecast errors of analysis fields generated by baselines under EDA using independent radiosonde observations, as shown in Figure \ref{fig7}. The forecasts driven by 4DVarFormer achieved error levels closest to the ERA5-driven forecasts across all evaluated variables, outperforming all others. Furthermore, although SDA closely matched the radiosonde observations in the analysis field error assessment, its forecast errors exhibited significant growth. This discrepancy indicates that the analysis fields generated by SDA struggle to satisfy physical consistency constraints, thereby compromising forecast accuracy.
	
	For the 10-day medium-range weather forecasting skill achieved with initial conditions produced by the deterministic DA configuration, please see the Supplementary Text as well as Figures S7 and S8.
	
	\FloatBarrier
	
	\FloatBarrier
	
	\subsection{Power Spectra of the analysis fields}
	
\begin{figure}[htbp]%
		\centering
		\includegraphics[width=0.9\textwidth]{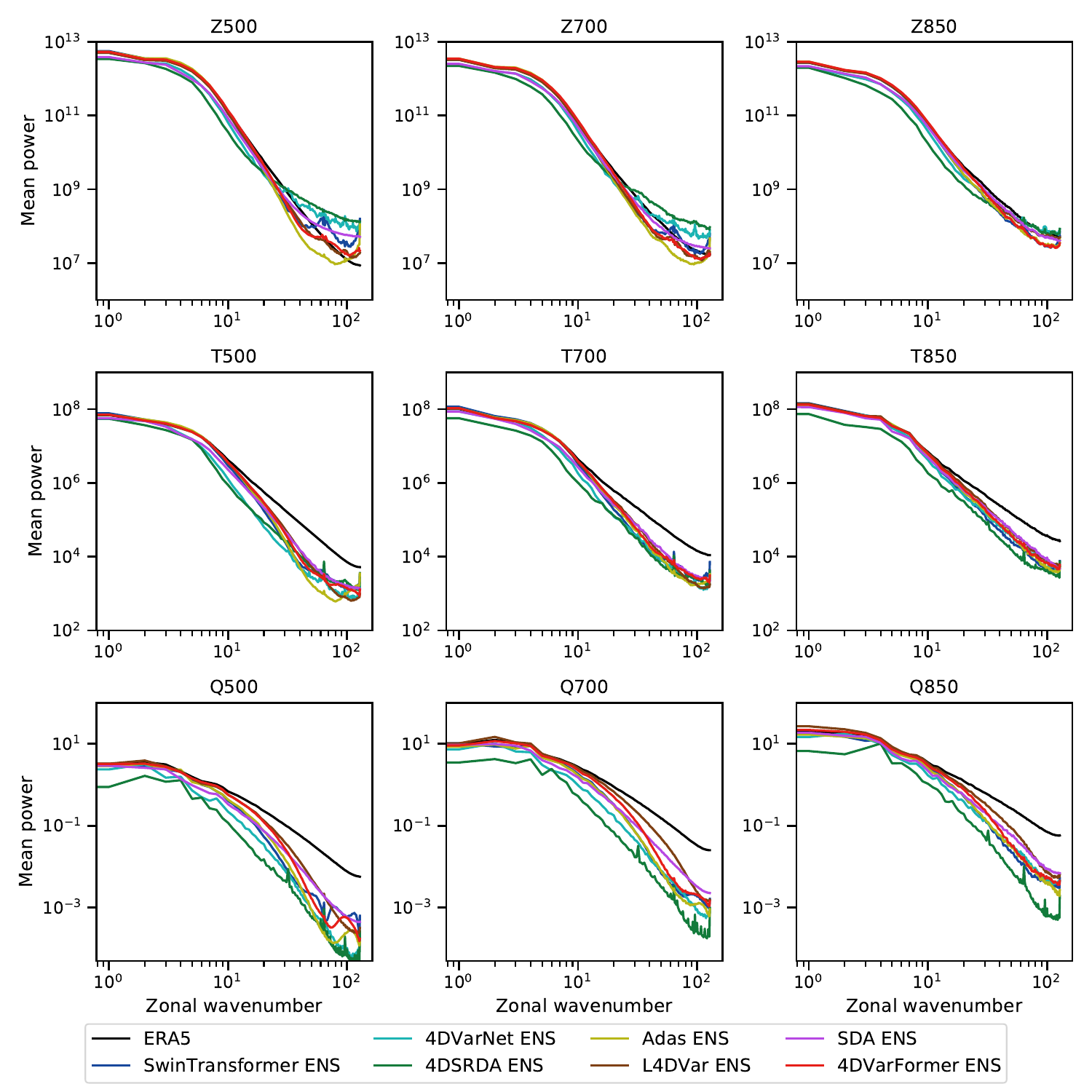}
		\caption{\textbf{Power spectra of the EDA geopotential, temperature, and specific humidity at 500 hPa, 700 hPa, and 850 hPa.} The power spectra of ERA5 are depicted by a black line. The power spectra of AI-based DA are color-coded as follows: SwinTransformer (dark blue), 4DVarNet (light blue), 4DSRDA (dark yellow-green), Adas (light yellow-green), L4DVar (brown), SDA (purple), and 4DVarFormer (red). These calculations are done for each day of the year at 00:00 UTC and 12:00 UTC. Each subplot represents a single variable, as indicated in the subplot titles.}
		\label{fig8}
	\end{figure}

	Figure~\ref{fig8} displays the power spectra of multiple variables for the AI-based DA baselines under the EDA configuration, using the ERA5 power spectra as the reference. In terms of Z500, the baselines 4DVarNet, 4DSRDA, and SDA all exhibit elevated energy levels at small scales, indicating the presence of high-frequency noise in their generated results. Conversely, Adas shows lower energy spectra compared to other baselines, suggesting that it obtains analysis field estimates by excessively smoothing atmospheric features. Furthermore, both 4DSRDA and 4DVarNet exhibit low energy spectra across all variables at large scales, indicating a deficiency in capturing large-scale atmospheric structures, which contributes to significant errors. The energy spectra of 4DVarFormer and L4DVar are relatively similar, suggesting that 4DVarFormer captures information across various scales with a fidelity comparable to L4DVar. Notably, current AI-based DA models still face challenges related to smoothing, particularly when employing the EDA configuration. For instance, the energy spectrum of SDA decreases from a level closest to ERA5 in deterministic DA to a level comparable to L4DVar and 4DVarFormer in the EDA configuration. This demonstrates that while ensemble methods can enhance DA stability, they may do so by smoothing out small-scale variability. For energy spectra results under the deterministic DA configuration, please refer to the Supplementary Text and Figure S9.
	
	\FloatBarrier
	
	\subsection{Visualization of the analysis fields}
	
\begin{figure}[htbp]%
		\centering
		\includegraphics[width=\textwidth]{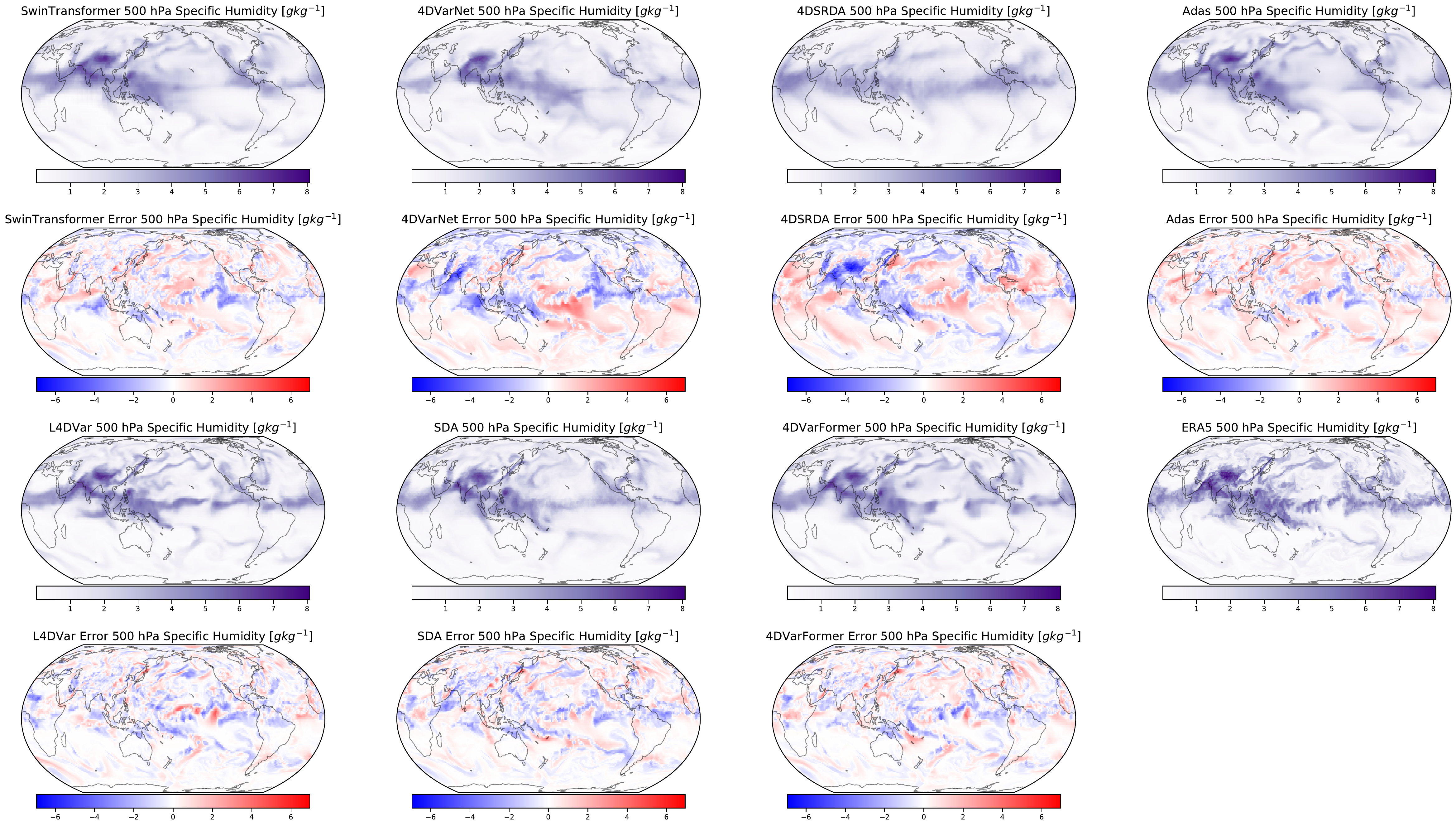}
		\caption{\textbf{Visualization of the 500 hPa specific humidity fields.} The analysis fields of baselines at 12:00 on July 23, 2023, are selected for visualization.}
		\label{fig9}
	\end{figure}

	Figure~\ref{fig9} presents the 500 hPa specific humidity analysis fields generated by various baseline methods under the EDA configuration, benchmarked against ERA5 data. Both 4DVarNet and 4DSRDA fail to accurately reconstruct the spatial structure of the humidity field, resulting in excessive smoothing. This outcome aligns with the low power spectra over all zonal wavenumbers observed in Figure~\ref{fig8}, indicating that these models even struggle to resolve large-scale details. Consequently, these models exhibit pronounced errors. In contrast, SwinTransformer and Adas capture the large-scale patterns but suffer from persistent blurring. This demonstrates that the direct fusion of observations with background fields is inadequate for representing the highly nonlinear humidity field. Although the SDA analysis field appears less blurred, it contains significant noise and yields larger overall errors than L4DVar and 4DVarFormer, particularly in mid-latitude regions. Among the evaluated baselines, L4DVar demonstrates superior spatial structure representation and lower errors. By incorporating 4DVar constraints, 4DVarFormer enhances the ability of AI-based DA models to capture nonlinearities, achieving performance comparable to L4DVar. Nevertheless, all baselines exhibit some degree of smoothing, underscoring the need for further improvements in resolving small-scale features within AI-based DA. Visualization results for additional variables are available in the Supplementary Text and Figures S10 to S22.
	
	\FloatBarrier
	
	\section{Discussion} 
	
	The rapid advancement of LWMs has sparked interest in employing these models to create operational data-driven global weather forecasting systems. Currently, LWMs rely on analysis or reanalysis fields generated by traditional NWP systems as inputs for their forecasts, which limits their autonomy as self-consistent and stand-alone systems. Recent research has demonstrated that AI-based DA models hold significant potential to enable operational AI-based weather forecasting. However, evaluating these models requires an objective, comprehensive benchmark based on real-world observations. To address this gap, we developed DABench by preprocessing ERA5 reanalysis alongside real-world GDAS prepbufr observations. This benchmark enhances our understanding of existing AI-based DA methods and provides a platform for fair and comprehensive comparison. Our results demonstrate that 4DVarFormer achieves superior robustness and accuracy compared to existing DA baselines, maintaining stable and skillful DA cycles for more than a year. Notably, 4DVarFormer achieves DA and medium-range weather forecasting metrics comparable to those of L4DVar, demonstrating the feasibility of developing advanced AI-based DA models for operational data-driven global weather forecasting systems.

	A limitation of this study is that we have not considered the assimilation of raw satellite observations, which constitute the primary data source for operational NWP systems~\cite{ECMWF_Doc_Obs}. Satellite observations provide critical information about atmospheric state, particularly in data-sparse regions such as oceans and polar areas, and offer high temporal and spatial coverage. Future research should focus on integrating satellite observations with AI-based DA methods, potentially through developing AI-based observation operators that can directly process satellite radiances. This would enable the development of more comprehensive DA models and create data-driven global weather forecasting systems that approach operational capabilities.
	
	The spatial resolution of 1.40625$^\circ$ in the current benchmark is sufficient for research purposes but falls short of the requirements for operational weather forecasting systems, which typically employ resolutions of 0.25$^\circ$ or higher~\cite{IFSUserGuide}. Increasing spatial resolution presents several challenges for AI-based DA methods. First, the exponential increase in computational cost and data storage requirements necessitates more efficient algorithms and hardware solutions. Second, higher resolution introduces smaller-scale atmospheric processes that may not be well-represented in current training datasets, potentially limiting model generalization. Third, the increased dimensionality exacerbates the curse of dimensionality in DA, making it more difficult to accurately estimate error covariances. Addressing these challenges will require innovations in model architecture, training strategies, and computational efficiency.
	
	We invite the meteorological and AI communities to participate in the development of novel AI-based DA models to accelerate the advancement of operational data-driven global weather forecasting systems. Building on our findings, we identify several promising research directions. First, developing ensemble-based AI DA methods that can provide uncertainty quantification without the computational cost of traditional ensemble DA. Second, exploring physics-informed neural networks that can incorporate dynamical constraints directly into the learning process, potentially improving physical consistency and reducing error accumulation. Third, investigating self-supervised and semi-supervised learning paradigms that can leverage the vast amount of unlabeled atmospheric data, reducing reliance on high-quality reanalysis products. Fourth, developing adaptive DA strategies that can dynamically adjust assimilation weights based on observation quality and forecast uncertainty, mimicking the quality control mechanisms in operational systems.
	
	Our research aims not to replace traditional NWP systems, but to complement and enhance them through AI-based innovations. Traditional NWP systems, refined over decades and rigorously evaluated in diverse real-world contexts, remain invaluable for operational weather forecasting and for providing high-quality datasets for AI. However, AI-based weather forecasting systems offer unique advantages in computational efficiency and the ability to learn complex patterns directly from data. Our findings demonstrate that AI-based DA models can effectively address the complexities inherent in real-world forecasting by autonomously assimilating observations. Looking ahead, we envision a hybrid paradigm where AI-based and physics-based methods synergistically combine to create more accurate, efficient, and scalable weather forecasting systems. The open-source nature of DABench and the emphasis on methodological reproducibility will facilitate this vision by enabling collaborative innovation across research teams.
	
	\section{Methods} \label{sec:4} 
	\subsection{General problem definition} \label{sec:4.1}
	This study considers a weather system that can be represented as follows:
	\begin{equation}
		\mathbf{x}(t_k) = \mathcal{M}_{t_{k-1} \rightarrow t_k}(\mathbf{x}(t_{k-1})),
	\end{equation}
	where $\mathbf{x}(t_k) \in \mathbb{R}^m$ denotes the system state at the $t_k$ moment, and $m$ denotes the dimension of the state space. $\mathcal{M}_{t_{k-1} \rightarrow t_k}: \mathbb{R}^m \mapsto \mathbb{R}^m$ corresponds to the real-world weather system, which maps the system state at the $t_{k-1}$ moment into the state at the $t_k$ moment. This study employs a neural network denoted as $\mathcal{N}^\mathcal{M}: \mathbb{R}^m \mapsto \mathbb{R}^m$ to approximate this system, which is trained on the ERA5 dataset. In discrete time, the observations $\mathbf{y}(t_k)$ can be denoted as follows:
	\begin{equation}
		\mathbf{y}(t_k) = \mathcal{H}(\mathbf{x}(t_k)) + \mathbf{\varepsilon}^o(t_k) ,
	\end{equation}
	where $\mathcal{H}: \mathbb{R}^m \mapsto \mathbb{R}^n$ denotes the observation operator and $n$ denotes the observation space's dimension. The observation operator $\mathcal{H}$ is utilized to observe a set of local points from the whole system. The observation error is expressed as a system-independent random error $\mathbf{\varepsilon}^o(t_k)$, mainly comprising instrumentation and representation errors. Assuming that the observation errors follow a Gaussian distribution, \emph{i.e.,} $\mathbf{\varepsilon}^o \sim \mathcal{N}(0,\mathbf{R})$, where $\mathbf{R}$ denotes the observation error covariance matrix ~\cite{frei2013mixture}. The background field often comes from short-range prediction and is defined as follows:
	\begin{equation}
		\mathbf{x}^b(t_k) = \mathcal{N}^\mathcal{M}_{t_{k-1} \rightarrow t_k}(\mathbf{x}^a(t_{k-1})),
	\end{equation}
	where $\mathbf{x}^a(t_{k-1})$, called the ``analysis field'', is obtained using a DA method. 
	
	In the Bayesian formulation, the initial field is estimated as the posterior distribution $p(\mathbf{x}|\mathbf{y})$ of the unknown state $\mathbf{x}$ conditioned on the observation $\mathbf{y}$, which can be obtained using Bayes' rule as follows:
	\begin{equation}
		p(\mathbf{x}|\mathbf{y}) = \frac{p(\mathbf{x},\mathbf{y})}{p(\mathbf{y})} = \frac{p(\mathbf{y}|\mathbf{x})p(\mathbf{x})}{p(\mathbf{y})},
	\end{equation}
	where $p(\mathbf{y})$ denotes the marginal probability and $p(\mathbf{x}|\mathbf{y})$ denotes the posterior probability. To better understand fundamental concepts and conventional DA methods, please refer to~\cite{carrassi2018data}.
	
	The AI-based DA task aims to develop a neural network that utilizes background fields and observations to generate accurate initial fields. In NWP systems, predictions typically occur daily at 0:00 and 12:00 UTC, limiting the observations available for assimilation in a short-term timeframe (e.g., $\geq -12:00$ and $\leq 00:00$ UTC). The AI-based DA model $\mathcal{N}^{DA}$ can be described as follows:
	\begin{equation}
		\mathcal{N}^{DA}(\mathbf{x}^b(t_0),\mathbf{y}): \mathbf{x}^b(t_0),\mathbf{y} \mapsto \mathbf{x}^n(t_0) ,
	\end{equation}
	where $\mathbf{x}^n_0$ denotes the neural network's assimilation result, $\mathbf{y}$ describes the observations in the DAW, and the goal is to make the neural network's output approach the reference system state $\mathbf{x}^t_0$ at the initial time $t_0$, i.e., $(\|{\mathbf{x}^n_0}-\mathbf{x}^t_0\| \sim 0)$. To prevent lookahead effects in AI-based DA models that compromise real-time analysis, DABench evaluates only the atmospheric state at the end of the DAW for models assimilating multiple observations within that period. Specifically, the analysis field at $-12:00$ UTC is propagated forward by the model to $00:00$ UTC, and this forecasted state is used as the evaluation target.
	
	\subsection{Datasets}~\label{sec:4.2}
	\subsubsection{Global weather reference} 
	\begin{table}[htbp]
		\caption{List of variables contained in the benchmark ground truth dataset. All fields have a dimensional $lat \times lon \times level$. The number of vertical levels for upper-air variables is given in the table, while the level of surface variables and constants is 1.}
		\label{tab1}
		\centering
		\footnotesize{\begin{tabular}{l|l}
				\hline
				Pressure level variables (Short name \& Unit) & Surface variables/Constants (Short name \& Unit)       \\
				\hline
				Geopotential (Z \& $m^2 s^{-2}$)  & 2m Temperature (T2M \& K)     \\
				Temperature (T \& K)  & 10m U Component of Wind (U10 \& $m s^{-1}$)     \\
				Specific Humidity (Q \& $kg kg^{-1}$)  & 10m V Component of Wind (V10 \& $m s^{-1}$)     \\
				U Component of Wind (U \& $m s^{-1}$)  & Mean Sea Level Pressure (MSLP \& $Pa$)     \\
				V Component of Wind (V \& $m s^{-1}$)  & land binary mask (lsm \& 0/1)     \\
				& Orography (orography \& m)\\
				\hline
		\end{tabular}}
	\end{table}
	\FloatBarrier
	
	The reanalysis dataset offers the most accurate historical weather state estimate at each time and location. Therefore, we utilized the ERA5 dataset~\cite{hersbach2020era5} as the global weather reference for training and testing the DA models. We chose hourly data ranging from 2010 to 2023 and conducted spatial interpolation from a $0.25^\circ$ latitude/longitude grid ($721 \times 1440$ grid points) to a $1.40625^\circ$ latitude/longitude grid ($128 \times 256$ grid points) following the data construction methodology used in WeatherBench~\cite{rasp2020weatherbench}. This approach reduces the I/O and memory load for model training. The interpolation was performed utilizing the bilinear interpolation algorithm in the xESMF Python library~\cite{jiawei_zhuang_2023_8356796}. Additionally, we selected 9 vertical levels in the upper air, including 50, 200, 250, 300, 500, 700, 850, 925, and 1000 hPa, which correspond to the pressure levels of the publicly available forecasts on the \href{https://confluence.ecmwf.int/display/TIGGE}{TIGGE archive}. We also focused on surface variables such as the 10m wind field, 2m temperature, and mean sea level pressure. The dataset is detailed in Table~\ref{tab1}, and stored in the NetCDF file format. We provide Python scripts to convert it to HDF5 for researchers to use, thus accelerating I/O during training.
	
	\subsubsection{Background field} The background fields employed for the training and validation of our DA model were derived from a 120-hour forecast based on the ERA5 initialization of Pangu-Weather~\cite{bi2023accurate}. During the final inference phase, the first background field is also a 120-hour forecast initialized from the ERA5 reanalysis, whereas the background fields in subsequent DA and forecast cycles are derived from forecasts initialized with analysis fields generated by the DA models, as shown in Figure~\ref{fig1} (b). The background fields are saved in the HDF5 file format.
	
	\subsubsection{Observations:} To evaluate the robustness of AI-based DA methods, we employed authentic observational datasets that encompass scenarios where certain instruments and satellite systems cease operation or are decommissioned. This approach enables AI-based DA models to develop proficiency in processing observations under complex conditions within a continuously evolving real-world atmospheric environment, thereby generating accurate analysis fields. Thus, this study utilizes real-world operational observational datasets for training and evaluating AI-based DA models. Here, we employ the GDAS prepbufr observations obtained from the National Centers for Environmental Prediction (NCEP) Automated Data Processing (ADP) global upper-air and surface weather observations available through the NCAR Research Data Archive (NCAR RDA)~\cite{cisl_rda_dsd337000}. These data primarily include conventional observations from land and marine stations, as well as satellite-derived data, with the latter constituting levels 2+ processed atmospheric products. The ground-based observations include land and marine surface reports, aircraft data, and radiosonde and pilot balloon observations. The satellite-based retrievals are supplied by the National Environmental Satellite Data and Information Service (NESDIS) and include oceanic wind data derived from the Special Sensor Microwave Imager (SSMI) and upper wind from Low Earth Orbit (LEO) and Geostationary Orbit (GEO) satellites. 
	
	\begin{figure}[htbp]%
		\centering
		\includegraphics[width=\textwidth]{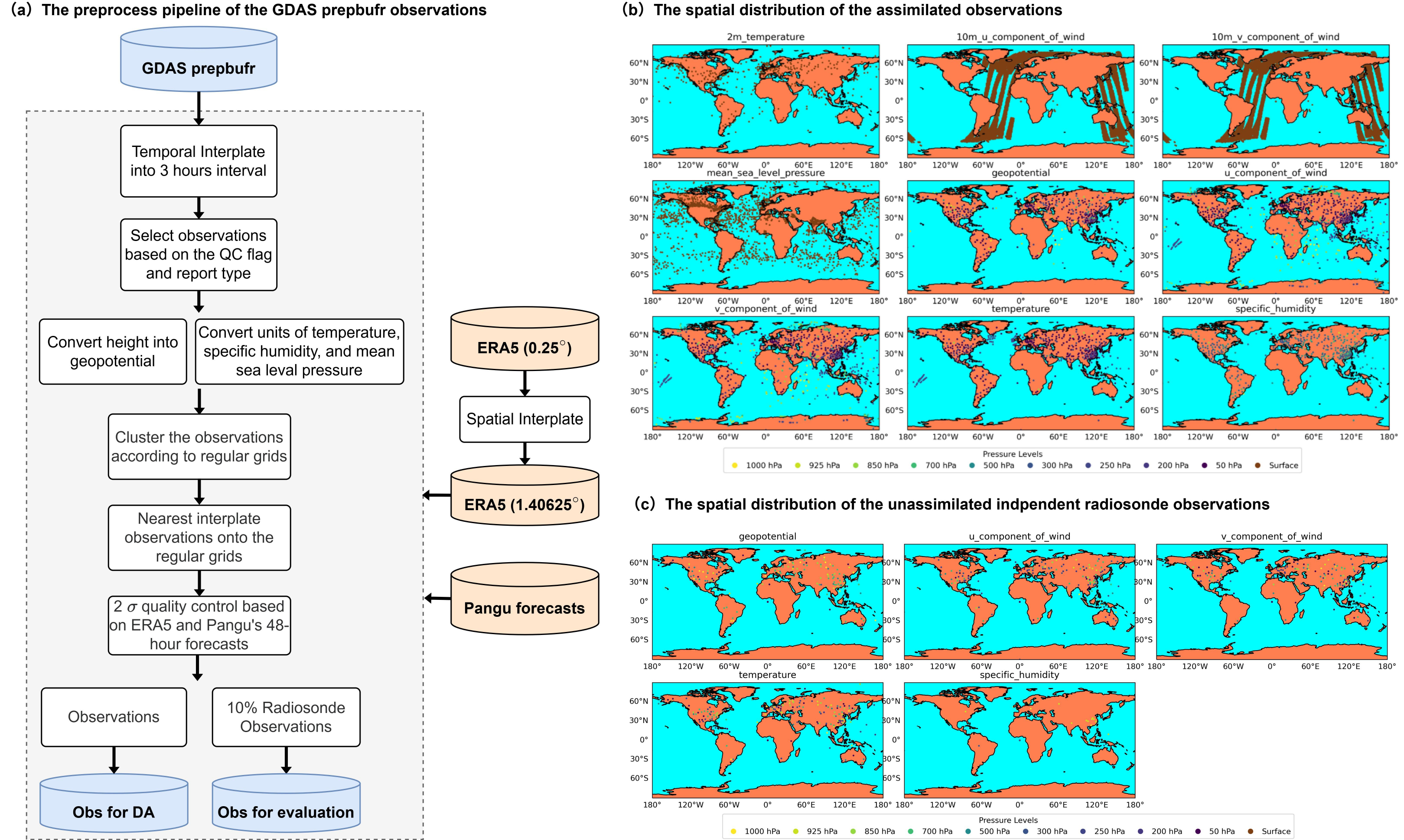}
		\caption{\textbf{Pre-process pipeline of the observations, as well as the spatial distribution of assimilated and independent observations.} Figure 10(a) represents the pre-process pipeline of the GDAS prepbufr observations. Figure 10(b) illustrates the spatial distribution of all observations assimilated at 00:00 UTC on February 1, 2023. In contrast, Figure 10(c) illustrates the spatial distribution of radiosonde observations that were not assimilated simultaneously, serving as independent verification data. Varying colors distinguish observations from different pressure levels.}
		\label{fig10}
	\end{figure}
	\FloatBarrier
	
	As shown in Figure~\ref{fig8}(a), we converted the GDAS prepbufr observations into 3-hour intervals and employed nearest-neighbor interpolation to map the data onto a grid aligned with the global weather reference, thereby achieving a spatial resolution of 1.40625$^\circ$. To maintain coherence with atmospheric variables conventionally employed in contemporary LWMs, we extracted the following observational variables from the GDAS prepbufr datasets corresponding to ``Report Type" classifications delineated in Table~\ref{tab2}, wherein the ``Note" designation explicitly specifies ``used by assimilation" within \href{https://www.emc.ncep.noaa.gov/mmb/data_processing/prepbufr.doc/table_2.htm}{the referenced table}: height, temperature, specific humidity, zonal wind velocity component, meridional wind velocity component, 2-meter temperature, 10-meter zonal wind velocity component, 10-meter meridional wind velocity component, and mean sea-level pressure. The geopotential observations underwent systematic conversion through implementation of the geometric transformation equation $z = \frac{g \times R_{\text{earth}} \times h}{R_{\text{earth}} + h}$, wherein $g$ represents the standard gravitational acceleration constant and $R_{\text{earth}}$ denotes the Earth's mean spherical radius. Furthermore, specific humidity magnitudes were subjected to multiplicative scaling by a scaling factor of $10e6$ to establish unit compatibility with the specific humidity unit employed within the ERA5 dataset. Subsequently, we calculated the distance (RMSE) for each observation with ERA5 serving as the reference dataset and removed observations where the distance exceeded twice the 2-day forecast error of Pangu-Weather for Quality Control (QC). We recalculated the statistical distance between the quality-controlled observations and ERA5 reanalysis and designated this metric as the observational error standard deviation for DA methods. The details of the observation error standard deviation values for each variable are shown in Table~\ref{tab3}. Please note that the designation ``NAN'' indicates that all observations of the corresponding variable were rejected during the QC process.
	
	\begin{table}[htbp]
		\caption{List of ``Report Type" selected for each variable.}
		\label{tab2}
		\centering
		\begin{tabular}{l|l}
			\hline
			Variable & Report Type \\
			\hline
			Z    & ``ADPUP", ``VADWND", ``PROFLR" \\
			U/V  & ``SYNDAT", ``ADPUP", ``VADWND", ``SFCSHP", ``ADPSFC", ``PROFLR",\\
			& ``AIRCFT", ``AIRCAR", ``SATWND", ``WDSATT", ``ASCATW", \\
			& ``ADPSFC, SFCSHP'' \\
			T    & ``ADPUP", ``RASSDA", ``AIRCFT", ``AIRCAR", ``SFCSHP", ``ADPSFC", \\
			&``ADPSFC, SFCSHP'' \\
			Q    & ``ADPUP", ``AIRCFT", ``SFCSHP", ``ADPSFC", ``ADPSFC, SFCSHP'' \\
			MSLP & ``SFCSHP", ``ADPSFC", ``ADPSFC, SFCSHP'' \\
			\hline
		\end{tabular}
	\end{table}
	
	\begin{table}[htbp]
		\caption{List of observation error standard deviation values for each variable.}
		\label{tab3}
		\centering
		\begin{tabular}{ccc|ccc}
			\hline
			Variable & Level & Value (Unit) & Variable & Level & Value (Unit) \\
			\hline
			Z & 50 hPa  & 432 ($m^2 s^{-2}$) & T & 50 hPa & 0.973 (K) \\
			Z & 200 hPa & 148 ($m^2 s^{-2}$) & T & 200 hPa & 0.748 (K) \\
			Z & 250 hPa & 146 ($m^2 s^{-2}$) & T & 250 hPa & 0.756 (K) \\
			Z & 300 hPa & 138 ($m^2 s^{-2}$) & T & 300 hPa & 0.626 (K) \\
			Z & 500 hPa & 97 ($m^2 s^{-2}$) & T & 500 hPa & 0.685 (K) \\
			Z & 700 hPa & 76 ($m^2 s^{-2}$) & T & 700 hPa & 0.913 (K) \\
			Z & 850 hPa & 77 ($m^2 s^{-2}$) & T & 850 hPa & 1.176 (K) \\
			Z & 925hPa & 76 ($m^2 s^{-2}$) & T & 925hPa & 1.197 (K) \\
			Z & 1000 hPa & 75 ($m^2 s^{-2}$) & T & 1000 hPa & 1.308 (K) \\
			\hline
			U & 50 hPa  & 1.827 ($m s^{-1}$) & V & 50 hPa & 1.778 ($m s^{-1}$) \\
			U & 200 hPa & 1.986 ($m s^{-1}$) & V & 200 hPa & 2.585 ($m s^{-1}$)  \\
			U & 250 hPa & 2.829 ($m s^{-1}$) & V & 250 hPa & 2.825 ($m s^{-1}$) \\
			U & 300 hPa & 2.900 ($m s^{-1}$) & V & 300 hPa & 2.853 ($m s^{-1}$) \\
			U & 500 hPa & 2.322 ($m s^{-1}$) & V & 500 hPa & 2.289 ($m s^{-1}$)  \\
			U & 700 hPa & 1.942 ($m s^{-1}$) & V & 700 hPa & 1.883 ($m s^{-1}$) 
			\\
			U & 850 hPa & 1.756 ($m s^{-1}$) & V & 850 hPa & 1.691 ($m s^{-1}$)   \\
			U & 925hPa & 1.734 ($m s^{-1}$) & V & 925hPa & 1.743 ($m s^{-1}$)   \\
			U & 1000 hPa & 1.498 ($m s^{-1}$) & V & 1000 hPa & 1.515 ($m s^{-1}$)   \\
			\hline
			Q & 50 hPa  & NAN ($g kg^{-1}$) & T2M & surface  & 1.225 (K) \\
			Q & 200 hPa  & NAN ($g kg^{-1}$) & U10 & surface  & 1.099 ($m s^{-1}$) \\
			Q & 250 hPa  & NAN ($g kg^{-1}$) & V10 & surface  & 1.152 ($m s^{-1}$) \\
			Q & 300 hPa  & 6.302e-5 ($g kg^{-1}$) & MSLP & surface & 101 (Pa) \\
			Q & 500 hPa & 3.135e-4 ($g kg^{-1}$) & & & \\
			Q & 700 hPa & 6.069e-4 ($g kg^{-1}$) & & & \\
			Q & 850 hPa & 7.773e-4 ($g kg^{-1}$) & & & \\
			Q & 925hPa & 6.900e-4 ($g kg^{-1}$) & & & \\
			Q & 1000 hPa  & 6.063e-4 ($g kg^{-1}$) & & & \\
			\hline
		\end{tabular}
	\end{table}
	
	Currently, most AI-based DA methods~\cite{chen2023adas,sun2025data,wang2024accurate} necessitate the interpolation of observations onto regularized spatial grids, thereby introducing systematic uncertainties and potential error propagation into the DA framework. In recognition of this inherent limitation, we preserved the original geographical coordinate information within the interpolated observational datasets, thereby serving standard data to facilitate future advancement of sophisticated DA techniques capable of incorporating observations with stochastic spatial distributions and irregular sampling geometries.
	
	\paragraph{Independent radiosondes reference:} To evaluate the fidelity of AI-based DA methods against the true atmospheric state, we distinguish between assimilated observations and independent radiosonde observations. Contemporary AI-based DA models predominantly rely on ERA5 reanalysis for training and validation~\cite{chen2023adas,huang2024diffda,sun2025data,wang2024accurate}, but reanalysis fields are inherently numerical model-constrained. Thus, direct comparison with raw observations is essential for the potential of AI-based DA methods to approximate real-world atmospheric states.
	
	To ensure the statistical independence of these reference data from the assimilated observations, the sampling was conducted on the radiosonde observations (BUFR code: ``ADPUP'') across each vertical pressure coordinate. Taking February 1, 2023, at 00:00 UTC as a representative case, Figure~\ref{fig8} (b) and Figure~\ref{fig8} (c) present the spatial distribution of assimilated observations alongside those radiosondes reserved for independent verification. The data reveal a predominant concentration of observations over terrestrial regions and their adjacent airspace. Notably, the 10-meter wind field measurements are chiefly sourced from satellite-derived products, whereas the mean sea-level pressure observations primarily originate from discrete platforms such as ship-based instruments. Within DABench, the observations have been systematically quantified across the training, validation, and test datasets. The results highlight the pronounced sparsity of observations, which poses considerable challenges for AI-driven data assimilation methodologies. Detailed statistics regarding the observation counts are provided in the Supplementary Tables S1 to S6.
	
	\subsection{Forecasting model} \label{sec:4.3}
	To rigorously validate the DA methods and assess the influence of the generated analysis fields on medium-range weather forecasting, we trained the Pangu-Weather model~\cite{bi2023accurate} using the collected ERA5 dataset. Detailed specifications of the hyperparameters are provided in the Supplementary Text.
	
	\subsection{DA baselines} \label{sec:4.4}
	
\begin{figure}[htbp]%
		\centering
		\includegraphics[width=\textwidth]{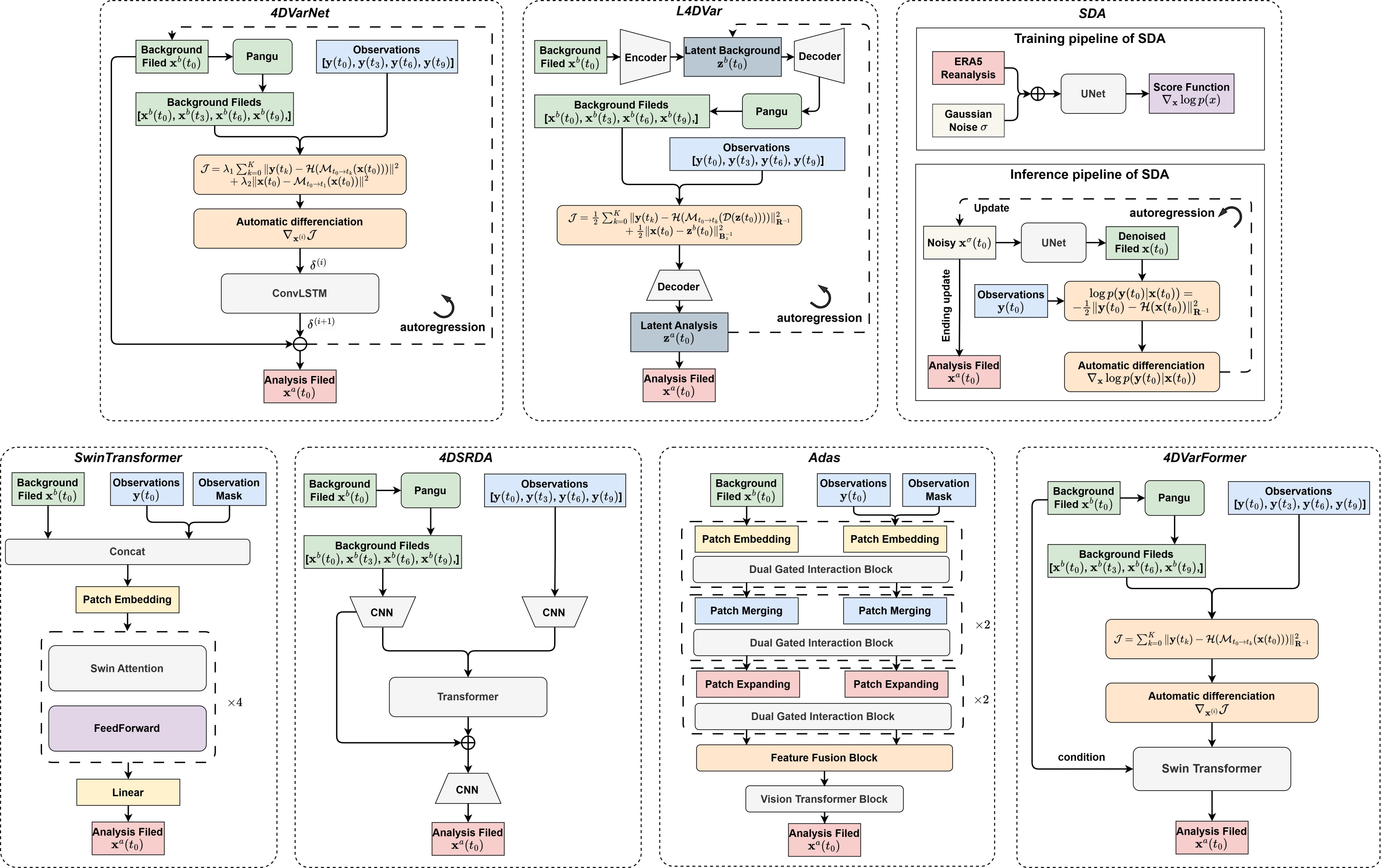}
		\caption{\textbf{Baselines evaluated in DABench.} We categorize baselines into two types: those requiring autoregressive iterations, such as 4DVarNet, L4DVar, and SDA; and those producing single-step output analysis fields, such as SwinTransformer, 4DSRDA, Adas, and 4DVarFormer.}
		\label{fig11}
	\end{figure}

	\begin{itemize}
		\item \textbf{SwinTransformer} As illustrated in Figure~\ref{fig11}, we establish SwinTransformer~\cite{liu2021swin} as a straightforward baseline, leveraging its ability to directly learn the mapping from background fields and observations to corresponding reanalysis fields. The background field and observations (including the mask matrix) are concatenated along the channel dimension, serving as inputs to the following four SwinTransformer blocks. Finally, the model generates the resulting analysis field.
		\item \textbf{4DVarNet} We transfer the SOTA 4DVarNet~\cite{fablet2021learning} model utilized in the domain of sea surface height reconstitution to our benchmark. The model architecture diagram is represented in Figure~\ref{fig11}. To adapt it to our task and dataset, we fixed the weights of the pre-trained forecast model during the 4DVarNet training process. This ensures that only the solver module for optimizing the 4DVar cost function is trained. The solver module utilizes a Convolutional Long Short-Term Memory (ConvLSTM) model to combine hidden features from historical iteration steps and outputs optimized increments. The 4DVar cost function to be optimized in the 4DVarNet model is as follows:
		\begin{equation}
			\mathcal{J}=\lambda_1\sum^K_{k=0}\|\mathbf{y}(t_k)-\mathcal{H}(\mathcal{N}^{\mathcal{M}}_{t_0 \rightarrow t_k}(\mathbf{x}^{(i)}(t_0)))\|^2+\lambda_2\|\mathbf{x}^{(i)}(t_{0})-\mathcal{N}^{\mathcal{M}}_{t_0 \rightarrow t_1}(\mathbf{x}^{(i)}(t_0))\|^2,
		\end{equation}
		where $\mathcal{N}^{\mathcal{M}}_{t_{k-1} \rightarrow t_{k}}$ denotes the pre-trained weather prediction model Pangu-Weather, $\mathbf{x}^{(i)}(t_0)$ represents the analysis field to be optimized at the $i$th iteration, $\mathbf{y}(t_k)$ denotes the observations at the $t_k$ moment, and $\mathcal{H}$ represents the observation operator. In this study, $\mathcal{H}$ is the mask matrix. Moreover, $\lambda_1$ and $\lambda_2$ represent the weight factors of the two components of the cost function.
		\item \textbf{4DSRDA} As illustrated in Figure~\ref{fig11}, we adapt the open-source model recently employed for flow field assimilation~\cite{yasuda2023spatio} to the task of weather forecasting. Specifically, it effectively maps the background field and observations within a DAW to a unified latent space. This process involves two downsampling blocks, each constructed by a convolutional neural network. Subsequently, it fuses these two features through a Transformer model and employs upsampling along with skip connections to produce the analysis field.
		\item \textbf{Adas} As illustrated in Figure~\ref{fig11}, we train the open-source model recently employed for global weather assimilation~\cite{chen2023adas} using DABench. Specifically, it utilizes the Transformer model to fuse the background field and observations by adopting a mask matrix to denote the importance of each observation. Here, we did not employ simulated observations sampled from ERA5 for pre-training. Instead, we directly utilized GDAS prepbufr data to train Adas. For comprehensive details regarding the model's architectural specifications, readers are directed to consult the original publication~\cite{chen2023adas}.
		\item \textbf{L4DVar} As illustrated in Figure~\ref{fig11}, L4DVar~\cite{fan2026physically} represents an SOTA AI-driven 4DVar framework that streamlines the DA process by replacing traditional tangent linear and adjoint models with the automatic differentiation capabilities of the LWM. Furthermore, it employs a diagonalized background error covariance matrix within a latent space, which significantly simplifies the implementation of the 4DVar algorithm without compromising DA accuracy. 
		\item \textbf{SDA} As illustrated in Figure~\ref{fig11}, Score-based Data Assimilation (SDA)~\cite{manshausen2025generative} is a DA approach grounded in diffusion models~\cite{yang2023diffusion}. Firstly, a score function is pre-trained on ERA5 reanalysis data to capture the underlying atmospheric manifold. Subsequently, during the inference phase, the gradient of an observation-based likelihood function is incorporated as guidance to reconstruct atmospheric fields from sparse observations.
		\item \textbf{4DVarFormer} As illustrated in Figure~\ref{fig11}, 4DVarFormer~\cite{wang2024accurate} is a Transformer model that integrates 4DVar a priori knowledge, effectively characterizing the relationships between wind-pressure relationship and temperature-humidity relationship. This model ensures accurate predictions of multivariate, three-dimensional atmospheric fields for the East China region within the simulated observations. We have adapted it for this benchmark to deal with the real-world global weather DA task.
		The 4DVar cost function used in 4DVarFormer is as follows:
		\begin{equation}
			\mathcal{J}=\frac{1}{2}\|\mathbf{x}(t_0)-\mathbf{x}^b(t_0)\|^2_{\mathbf{B}^{-1}} + \frac{1}{2}\sum^K_{k=0}\|\mathbf{y}(t_k)-\mathcal{H}(\mathcal{N}^{\mathcal{M}}_{t_0 \rightarrow t_{k}}(\mathbf{x}(t_0)))\|^2_{\mathbf{R}^{-1}}.
		\end{equation}
		Since our model is non-autoregressive, we only need to input $\mathbf{x}(t_0)=\mathbf{x}^b(t_0)$ and calculate the derivative. In this context, the first term of the above cost function is equal to 0. Therefore, we can use the following simplified cost function:
		\begin{equation}
			\mathcal{J}=\sum^K_{k=0}\|\mathbf{y}(t_k)-\mathcal{H}(\mathcal{N}^{\mathcal{M}}_{t_0 \rightarrow t_{k}}(\mathbf{x}^b(t_0)))\|^2_{\mathbf{R}^{-1}}.
		\end{equation}
		Here $\mathbf{R}$ is represented as follows:
		\begin{equation}
			\mathbf{R} = 
			\begin{bmatrix}
				\mathbf{R}_{T2M} \\
				& \mathbf{R}_{U10} \\
				& & \mathbf{R}_{V10} \\
				& & & \mathbf{R}_{MSLP} \\
				& & & & \mathbf{R}_{Z} \\
				& & & & & \mathbf{R}_{U} \\
				& & & & & & \mathbf{R}_{V} \\
				& & & & & & & \mathbf{R}_{T} \\
				& & & & & & & & \mathbf{R}_{Q} \\
			\end{bmatrix},
		\end{equation}
		where
		\begin{align}
			\mathbf{R}_{T2M} &= \sigma^2_{T2M}\mathbf{I}_{HW \times HW}, \\
			\mathbf{R}_{U10} &= \sigma^2_{U10}\mathbf{I}_{HW \times HW}, \\
			\mathbf{R}_{V10} &= \sigma^2_{V10}\mathbf{I}_{HW \times HW}, \\
			\mathbf{R}_{MSLP} &= \sigma^2_{MSLP}\mathbf{I}_{HW \times HW}, \\
			\mathbf{R}_{Z} &= \sigma^2_{Z}\mathbf{I}_{9HW \times 9HW}, \\
			\mathbf{R}_{U} &= \sigma^2_{U}\mathbf{I}_{9HW \times 9HW}, \\
			\mathbf{R}_{V} &= \sigma^2_{V}\mathbf{I}_{9HW \times 9HW}, \\
			\mathbf{R}_{T} &= \sigma^2_{T}\mathbf{I}_{9HW \times 9HW}, \\
			\mathbf{R}_{Q} &= \sigma^2_{Q}\mathbf{I}_{9HW \times 9HW}, \\
		\end{align}
		and $\mathbf{R}_{X}$ represents the observations error covariance matrix of variable $X$. $\sigma_{X}$ denotes the error standard deviation of the variable $X$.
	\end{itemize}
	
	\FloatBarrier
	
	The baseline models were trained and tested using the dataset described in Section~\ref{sec:4.2}. The training set consisted of data from 2010 to 2021. The year 2022 data was used as the validation dataset. Finally, the year 2023 data was used to test all models. All models were trained using background fields generated from a 120-hour prediction initialized by ERA5 data. The training process for all models was conducted using 4 NVIDIA 3090 GPUs. Please refer to the Supplementary Text for the training details of these models above.
	
	\subsection{EDA configuration} \label{sec:4.5}
	We generated an 11-member ensemble for the DA cycle. Consistent with the methodology used by ECMWF for ensemble simulations~\cite{buizza1999stochastic}, which involves perturbing both initial conditions and model physics, we introduced 10 sets of random Perlin noise~\cite{bi2023accurate} into the background field at the beginning of the DA cycle. Additionally, we applied Monte Carlo dropout (MC dropout)~\cite{gal2016dropout} with a dropout rate of 0.2 to perturb the DA and forecasting model parameters. The perturbation method, similar to that used in Pangu-Weather~\cite{bi2023accurate}, generates perturbations using 3 octaves of Perlin noise with scales of 0.2, 0.1, and 0.05, and the number of periods along each axis (longitude and latitude) is set to 1, 2, and 4, respectively. After the initial DA step, no further perturbations are added to the subsequent background fields, as the system autonomously generates EDA results for each time step.
	
	\subsection{Metrics} \label{sec:4.6}
	Our objective is to assess the performance of the DA baselines in accordance with the standard evaluation practices of the NWP systems. Consequently, the benchmark established in this study involves a comprehensive evaluation of a one-year DA cycle and a 10-day medium-range weather forecast. It is noteworthy that the ERA5 dataset employs a 12-hour DAW for its reanalysis fields generated at 00:00 and 12:00 each day~\cite{hersbach2020era5}. Moreover, due to the extremely sparse nature of GDAS prepbufr observations, achieving a stable DA and forecasting cycle is quite challenging in these conditions. Consequently, the DA cycle is conducted at 12-hour intervals to maximize the number of observations assimilated, thereby enhancing the stability of the DA process. For the SwinTransformer~\cite{liu2021swin}, Adas~\cite{chen2023adas}, and SDA~\cite{manshausen2025generative} models, we employ a 3-hour DAW, as there is no requirement to incorporate temporal observations. This approach allows the models to assimilate observations exclusively at the analysis time to construct the analysis field. Furthermore, since our observations consider data within 1.5 hours before and after the target time as equivalent, SwinTransformer~\cite{liu2021swin}, Adas~\cite{chen2023adas}, and SDA~\cite{manshausen2025generative} all have a lookahead of 1.5 hours. In contrast, for the 4DVarNet~\cite{fablet2021learning}, 4DSRDA~\cite{yasuda2023spatio}, L4DVar~\cite{fan2026physically}, and 4DVarFormer~\cite{wang2024accurate} models, we assimilate observations from the [-12 hours, -9 hours, -6 hours, -3 hours] interval to generate the analysis field at -12 hours. Subsequently, we utilize the Pangu-Weather~\cite{bi2023accurate} model to forecast from -12 hours to 0 hours, thereby obtaining the evaluation ``analysis field." This method is analogous to the approach used in IFS for deriving the 0-hour field, but our configuration eliminates any lookahead. Consequently, the lookahead time for our AI-based DA baselines reaches a maximum of 1.5 hours, marking a significant enhancement compared to ERA5's 9 hours~\cite{hersbach2020era5}. Specifically, the DA cycle is run for the year 2023 at 00:00 UTC and 12:00 UTC each day, which corresponds to the initialization times for the 10-day forecast conducted by the IFS High-RESolution (HRES), Pangu-Weather~\cite{bi2023accurate}, GraphCast~\cite{lam2023learning}, and FengWu~\cite{chen2023fengwu}. The performance of 10-day medium-range weather forecasts is evaluated across the 2023 annual cycle, utilizing initial fields at 00:00 UTC and 12:00 UTC, with a time integral of 6-hours.
	
	All metrics were computed using float32 precision and reported using the native scale of the variables without normalization. Notably, all metrics are computed using a latitude-weighting factor over grid points due to the non-equal area distribution from the equator towards the north and south poles. Let $\alpha_j$ be the latitude weighting factor for the latitude at the $j$th latitude index, which is defined as
	\begin{equation}
		\alpha_j = \frac{\cos{lat(j)}}{\frac{1}{H}\sum^{H}_{\hat{j}}\cos{lat(\hat{j})}},
	\end{equation}
	where $lat(j)$ represents the latitude of the $j$th grid, $H$ is the number of latitudes in a given resolution.
	
	\textbf{Weighted Root Mean Square Error (WRMSE)} We evaluate assimilate and forecast skill for a given variable, $\mathbf{x}$, compared to the ERA5 reanalysis field using a latitude-Weighted Root Mean Square Error (WRMSE)~\cite{rasp2020weatherbench} given by
	\begin{equation}
		\texttt{WRMSE} = \frac{1}{|D_{eval}|}\sum_{i \in D_{eval}}\sqrt{\frac{1}{HW}\sum^{H}_j\sum^{W}_k \alpha_j(\hat{\mathbf{x}}_{i,j,k}-\mathbf{x}^t_{i,j,k})^2},
	\end{equation}
	where 
	\begin{itemize}
		\item $\hat{\mathbf{x}}$ is the field to be evaluated,
		\item $\mathbf{x}^t$ is the ERA5 ground truth,
		\item $i \in D_{eval}$ represents the sample index in the evaluation dataset,
		\item $j$ represents the latitude coordinate in the grid,
		\item $k$ represents the longitude coordinate in the grid.
	\end{itemize}
	The lower the WRMSE, the better the results.
	
	\textbf{Observational Root Mean Square Error (ORMSE)} To assess assimilation performance for a specified variable $\mathbf{x}$ against the sparse radiosonde observations that remained unassimilated, we employ an Observational Root Mean Square Error (ORMSE) metric, which is formulated as follows:
	\begin{equation}
		\texttt{ORMSE} = \frac{1}{|D_{eval}|}\sum_{i \in D_{eval}}\sqrt{\frac{1}{N_{lat,lon}}\sum^{N_{lat,lon}}_{j,k} (\hat{\mathbf{x}}_{i,j,k}-\mathbf{x}^o_{i,j,k})^2},
	\end{equation}
	where
	\begin{itemize}
		\item $\mathbf{x}^o$ is the GDAS prepbufr observations that were not assimilated,
		\item $N_{lat,lon}$ represents the number of the observation locations.
	\end{itemize}
	The lower the ORMSE, the better the results.
	
	\textbf{Weighted Bias (WBias)} We also computed the latitude-Weighted Bias (WBias) for a given variable, $\mathbf{x}$
	\begin{equation}
		\texttt{WBias} = \frac{1}{|D_{eval}|}\sum_{i \in D_{eval}}\frac{1}{HW}\sum^{H}_j\sum^{W}_k \alpha_j(\hat{\mathbf{x}}_{i,j,k}-\mathbf{x}^t_{i,j,k}).
	\end{equation}
	The closer the WBias is to 0, the better the results are.
	
	\textbf{Observational Bias (OBias)} To evaluate the bias of the AI-based DA baselines against the radiosonde observations that remained unassimilated, we employ an Observational Bias (OBias) metric, which is formulated as follows:
	\begin{equation}
		\texttt{OBias} = \frac{1}{N_{lat,lon}}\sum^{N_{lat,lon}}_{j,k}(\hat{\mathbf{x}}_{i,j,k}-\mathbf{x}^o_{i,j,k}).
	\end{equation}
	The closer the OBias is to 0, the better the results are.
	
	\textbf{Anomaly Correlation Coefficient (ACC)} To study skillful forecast lead times, we also calculated the latitude-weighted Anomaly Correlation Coefficient (ACC)~\cite{rasp2020weatherbench} according to
	\begin{equation}
		\texttt{ACC} = \frac{1}{|D_{eval}|}\sum_{i \in D_{eval}}\frac{\sum^{H,W}_{j,k}\alpha_j\left(\hat{\mathbf{x}}_{i,j,k}-C_{j,k}\right)\left(\mathbf{x}^t_{i,j,k}-C_{j,k}\right)}{\sqrt{\left[\sum^{H,W}_{j,k}\alpha_j\left(\hat{\mathbf{x}}_{i,j,k}-C_{j,k}\right)^2\right]\left[\sum^{H,W}_{j,k}\alpha_j\left(\mathbf{x}^t_{i,j,k}-C_{j,k}\right)^2\right]}},
	\end{equation}
	where $C_{j,k}$ denotes the climatological mean for a given variable and the day-of-year containing the validity time. It is calculated referring to GraphCast~\cite{lam2023learning} and FengWu~\cite{chen2023fengwu}. The climatological mean was computed using ERA5 data between 2010 and 2021. The higher the ACC, the better the results.
	
	\textbf{Activity} To evaluate the smoothness of the forecasts, we introduce the activity metric according to
	\begin{equation}
		\texttt{Activity} = \frac{1}{|D_{eval}|}\sum_{i \in D_{eval}}\sqrt{\frac{1}{HW}\sum^{H,W}_{j,k} \alpha_j\left[(\hat{\mathbf{x}}_{i,j,k}-\mathbf{x}^t_{i,j,k})-\frac{1}{HW}\sum^{H,W}_{j,k} \alpha_j(\hat{\mathbf{x}}_{i,j,k}-\mathbf{x}^t_{i,j,k})\right]^2}.
	\end{equation}
	The lower the forecast activity, the smoother the forecast.

	\textbf{Power Spectra} The power spectra is computed for each pressure level by applying a Discrete Fourier Transform (DFT) to zonal cross-sections of atmospheric variables. For each latitude circle $j$, we extract the longitudinal profile $\mathbf{x}_{\text{all},j,k}$ where $k$ indexes the longitude dimension. The power spectrum is computed as:
	
	\begin{equation}
		S[m] = \frac{1}{N_j} \left| \sum_{k=0}^{N_j-1} \mathbf{x}_{\text{all},j,k} e^{-i 2\pi n_w k / N_j} \right|^2,
	\end{equation}
	where $N_j$ is the number of longitude points at latitude $j$, and $n_w$ is the zonal wavenumber. For zonally averaged component (m=0):
	\begin{equation}
		S[0] = \frac{1}{N_j} \left| \sum_{k=0}^{N_j-1} \mathbf{x}_{\text{all},j,k} \right|^2.
	\end{equation}
	
	To account for the varying grid cell areas with latitude, we apply a cosine-weighting correction:
	\begin{equation}
		S_{\text{corrected}}[n_w] = S[n_w] \cdot \cos(\text{latitude}[j]),
	\end{equation}
	where $\text{latitude}[j]$ is in radians. Finally, the global power spectrum is obtained by averaging across all latitude circles:
	\begin{equation}
		S_{\text{global}}[n_w] = \frac{\sum_j S_{\text{corrected}}[n_w,j] \cdot \cos^2(\text{latitude}[j])}{\sum_j \cos^2(\text{latitude}[j])}.
	\end{equation}
	
	This approach properly accounts for the spherical geometry and ensures that the power spectra reflect true spatial variability rather than geometric artifacts.
	
	\textbf{Continuous Ranked Probability Score (CRPS)} To assess the divergence of AI-based DA methods in our simple ensemble method, we evaluated the Continuous Ranked Probability Score (CRPS) of DA baselines in our EDA experiments. The CRPS was computed using the following equation:
	\begin{equation}
		\texttt{CRPS} = \int^{\infty}_{-\infty}\left[F(\hat{\mathbf{x}}_{j,k})-\mathcal{G}(\mathbf{x}^t_{j,k}\leq z)\right]dz,
	\end{equation}
	where $F(\hat{\mathbf{x}}_{j,k})$ represents the cumulative distribution function (CDF) of the $\hat{\mathbf{x}}_{j,k}$, and $\mathcal{G}$ is an indicator function. The indicator function equals $1$ if the statement $\mathbf{x}^t_{j,k} \leq z$ is true; otherwise, it takes the value of $0$. This study uses the xskillscore Python package to calculate the CRPS metric.
	
	The lower the CRPS, the better the results.
	
	\textbf{Spread-Skill Ratio (SSR)} The Spread-Skill Ratio (SSR) is defined as the ratio between the ensemble spread and the RMSE of the ensemble mean, where the following equation calculates the spread:
	\begin{equation}
		\texttt{Spread} = \sqrt{\frac{1}{HW}\sum^{H,W}_{j,k}\alpha_j \texttt{var}_m(\mathbf{X}_{j,k})},
	\end{equation}
	with $\texttt{var}_m$ being the variance in the ensemble dimension. Thus, we define SSR as follows:
	\begin{equation}
		\texttt{SSR} = \frac{\texttt{Spread}}{\texttt{RMSE}(\overline{\mathbf{X}})}.
	\end{equation}
	A smaller SSR indicates an underdispersive forecast, whereas a larger SSR indicates an overdispersive forecast. An SSR closer to 1 indicates that the ensemble dispersion is reasonably consistent with the forecast error, reflecting an appropriate level of ensemble spread.

	\backmatter
	
	\bmhead{Supplementary information}
	The supplementary material is available at the ``supplementary\_information.pdf".
	
	\bmhead{Acknowledgements}
	The authors extend their gratitude to the ECMWF and NCEP for their significant efforts to store and provide invaluable data, which are crucial for this work and the research community. Additionally, this work was carried out at the National Supercomputer Center in Tianjin, and the calculations were performed on Tianhe new generation supercomputer. We would also like to express our appreciation to the research team and service team in the Shanghai Artificial Intelligence Laboratory for the provision of computational resources and infrastructure. 
	
	\bmhead{Funding}
	Not applicable. 
	
	\bmhead{Conflict of interest}
	All authors declare no financial or non-financial competing interests. 
	
	\bmhead{Ethics approval and consent to participate}
	Not applicable. 
	
	\bmhead{Consent for publication}
	Not applicable. 
	
	\bmhead{Availability of data and materials}
	All data needed to evaluate the conclusions in the paper are present in the paper and/or the Supplementary Information. The DABench dataset is available at \href{https://pan.baidu.com/s/1H3_H2Xvy8OAZXIL-W7woYw?pwd=pdvn}{the Baidu Drive}. 
	
	\bmhead{Code availability}
	The source code used for the benchmark proposed in this work is available in a Github repository \href{https://github.com/wuxinwang1997/DABench}{https://github.com/wuxinwang1997/DABench}. The xskillscore Python package can be accessed from \href{https:// github.com/xarray-contrib/xskillscore/}{https:// github.com/xarray-contrib/xskillscore/}. The implementation of Perlin noise is based on publicly available code from the GitHub repository: \href{https://github.com/pvigier/perlinnumpy/}{https://github.com/pvigier/perlinnumpy/}.
	
	\bmhead{Author contribution}
	K.J.R., L.B., and W.X.W. designed the project. K.J.R., L.B., W.C.N., and B.H.D. managed and oversaw the project. W.X.W. and T.K.Y. performed the model training and evaluation. W.X.W. and T.H. improved the model design. W.X.W., T.K.Y., X.Y.L., and T.H. analyzed the experiment results. W.X.W., B.F., X.Y.L., and W.C.N. wrote and revised the manuscript.

	\bibliography{sn-bibliography}

\end{document}